\documentclass[11pt]{article}
\usepackage[utf8]{inputenc}
\usepackage{amsfonts}
\usepackage{booktabs}
\usepackage[preprint]{acl}

\usepackage{bbm}
\usepackage{times}
\usepackage{latexsym}
\usepackage{listings}
\usepackage{multirow}
\usepackage[T1]{fontenc}
\usepackage[utf8]{inputenc}
\usepackage{amsmath}
\usepackage{microtype}

\usepackage{inconsolata}

\usepackage{graphicx}

\title{MemBuilder: Reinforcing LLMs for Long-Term Memory Construction via Attributed Dense Rewards}

\author{
  \textbf{Zhiyu Shen}$^{1}$\thanks{Equal contribution.}\thanks{Work done during internship at Tencent Inc.}, 
  \textbf{Ziming Wu}$^{2*}$\thanks{Corresponding author.}, 
  \textbf{Fuming Lai}$^{2}$, 
  \textbf{Shaobing Lian}$^{2}$, 
  \textbf{Yanghui Rao}$^{1}$ \\
  $^{1}$School of Computer Science and Engineering, Sun Yat-sen University, Guangzhou, China \\
  $^{2}$Tencent Inc. \\
  \texttt{shenzhy23@mail2.sysu.edu.cn, raoyangh@mail.sysu.edu.cn}\\
  \texttt{\{jimmyzmwu, fuminglai, lokilian\}@tencent.com}
}

\begin{document}
\maketitle

\begin{abstract}
Maintaining consistency in long-term dialogues remains a fundamental challenge for LLMs, as standard retrieval mechanisms often fail to capture the temporal evolution of historical states.
While memory-augmented frameworks offer a structured alternative, current systems rely on static prompting of closed-source models or suffer from ineffective training paradigms with sparse rewards. 
We introduce MemBuilder, a reinforcement learning framework that trains models to orchestrate multi-dimensional memory construction with attributed dense rewards. 
MemBuilder addresses two key challenges: (1) Sparse Trajectory-Level Rewards: we employ synthetic session-level question generation to provide dense intermediate rewards across extended trajectories; and (2) Multi-Dimensional Memory Attribution: we introduce contribution-aware gradient weighting that scales policy updates based on each component's downstream impact.
Experimental results show that MemBuilder enables a 4B-parameter model to outperform state-of-the-art closed-source baselines, exhibiting strong generalization across long-term dialogue benchmarks.\footnote{Code available at \url{https://github.com/Zh1yuShen/MemBuilder}}
\end{abstract}

\section{Introduction}

Memory-augmented frameworks have emerged as a promising approach for maintaining consistency in long-term dialogues, which must track evolving contexts and historical states over extended timelines.
While Retrieval-Augmented Generation (RAG) facilitates access to external knowledge, it treats retrieval units as independent, static chunks—failing to capture how information evolves or which historical facts have been superseded \citep{liu2024lost, gao2024retrievalaugmentedgenerationlargelanguage}.
%While Retrieval-Augmented Generation (RAG) enables access to external knowledge, it treats each retrieval independently—returning isolated text chunks that lack awareness of how information evolves over time or which facts supersede earlier ones \citep{liu2024lost, gao2024retrievalaugmentedgenerationlargelanguage}.
Memory-augmented frameworks address this by decomposing information prior to storage: events receive independent timestamps and semantic concepts are structured into discrete units. 
This shifts the computational burden from processing entangled tokens during inference to retrieving precise, ``pre-digested" fragments. 
Recent implementations like Mem0 \cite{DBLP:journals/corr/abs-2504-19413}, MIRIX \cite{DBLP:journals/corr/abs-2507-07957}, and MemGPT \cite{DBLP:journals/corr/abs-2310-08560} exemplify this approach, constructing external memory that evolves with each interaction. 
However, these prompting-based systems face three key limitations: they operate in an ``open loop'' without feedback on whether the constructed memories benefit downstream tasks; their performance degrades sharply when substituting less capable models; and their reliance on expensive closed-source models for the token-intensive memory construction phase makes deployment cost prohibitive at scale.
This raises a critical question: \textbf{Can we instead train a model to perform memory construction through direct supervision?}
\begin{figure}[t]
    \centering
    \includegraphics[width=\linewidth, page=1]{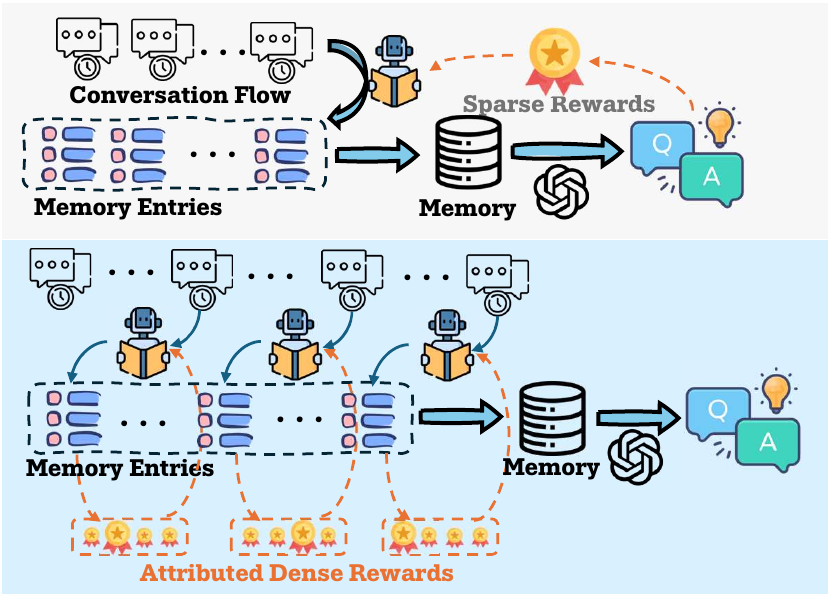}
    \caption{Sparse trajectory-level rewards (top) vs. our attributed dense session-level rewards (bottom). Dense rewards provide learning signals at each session rather than only at trajectory end.}
    \label{fig:intro}
\end{figure}

Current training-based approaches, such as Memory-R1 \citep{DBLP:journals/corr/abs-2508-19828} and Mem-$\alpha$ \citep{DBLP:journals/corr/abs-2509-25911}, attempt to address this learning gap but face two critical technical bottlenecks:
1) Sparse Trajectory-Level Rewards: In long-term dialogues, a single reward given at the end of a multi-session trajectory is too sparse. The model cannot discern which session's memory operations contributed to the final outcome, making gradient updates noisy and learning ineffective.
2) Multi-Dimensional Memory Attribution: While the previous method adopts a multi-dimensional 
memory, all components share a global reward, failing to distinguish between operations on different types of memories, regardless of actual downstream impact.
%While multi-dimensional memory architectures can offer finer-grained control \citep{DBLP:journals/corr/abs-2507-07957}, lightweight LLMs often violate the required action schema and constraints, producing invalid actions. As a result, direct RL sampling is dominated by non-executable rollouts, making optimization unstable and inefficient.

We present MemBuilder, a framework for training models to construct long-term memory with attributed dense rewards. Our architecture utilizes a multi-dimensional memory design, comprising Core, Episodic, Semantic and Procedural components, and trains a single, lightweight 4B model to manage all four components.
Our approach introduces two key technical contributions to address the limitations of existing RL-based memory construction methods (Figure~\ref{fig:intro}).
First, we employ \textbf{dense session-level rewards}. Unlike traditional methods that assign a single reward after processing all sessions, we leverage synthetic session-level question generation to provide immediate feedback after each session's memory operations.
Second, we introduce \textbf{contribution-aware gradient weighting} to resolve multi-dimensional memory attribution. Since all four memory components share a common reward, their individual contributions remain ambiguous. %Our mechanism dynamically amplifies gradients for memory operations based on the downstream utility of constructed memories. The more utilization of a specific memory component leads to the more impact of the corresponding operations on the training.
Our mechanism addresses this by scaling gradient updates based on the downstream utility of the constructed memories; specifically, the gradient impact of a memory operation is proportional to the usage of its corresponding memory component during retrieval.
We integrate these techniques into \textbf{Attributed Dense Rewards Policy Optimization (ADRPO)}.
Extensive experiments conducted on different benchmarks demonstrate the effectiveness of our proposed framework. Specifically, MemBuilder achieves 84.23\% on LoCoMo, surpassing the baselines including Claude 4.5 Sonnet under the same setting. Notably, our model is trained exclusively on LongMemEval yet generalizes effectively to out-of-distribution benchmarks with different dialogue structures and question types.

\section{Related Works}

\subsection{Long-Term Memory Management in Conversational Agents}

Maintaining coherent and personalized interactions over extended dialogues remains a fundamental challenge for LLM-based agents~\citep{DBLP:conf/aaai/ZhongGGYW24, DBLP:journals/corr/abs-2310-08560}. Recent benchmarks such as LoCoMo~\citep{DBLP:conf/acl/MaharanaLTBBF24}, LongMemEval~\citep{DBLP:conf/iclr/WuWYZCY25}, and PerLTQA~\citep{du-etal-2024-perltqa} evaluate long-term memory through multi-session QA, temporal reasoning, and evolving user profiles. Early approaches addressed context limitations through position encoding modifications~\citep{DBLP:journals/corr/abs-2309-12307, peng2023yarn} or dedicated long-context training~\citep{DBLP:conf/nips/TworkowskiSPWMM23, bai2024longalign}, but these incur high computational costs and struggle to capture the temporal dynamics in multi-session dialogues~\citep{liu2024lost}. While RAG offers better scalability \citep{gao2024retrievalaugmentedgenerationlargelanguage}, its chunk-level retrieval lacks the temporal and semantic organization needed for complex long-term reasoning \citep{liu2024lost}, motivating memory-augmented solutions that decompose information into structured units.

\subsection{Prompting-Based Memory Frameworks}

Inspired by the cognitive science distinctions among episodic, semantic, and procedural memory~\citep{tulving1972episodic}, recent frameworks construct structured external memory for LLM agents~\citep{sumers2024coala, laird2012soar}. Representative prompting-based implementations include MemGPT with its operating system-like memory hierarchy~\citep{DBLP:journals/corr/abs-2310-08560}, Mem0 for personalized memory extraction~\citep{DBLP:journals/corr/abs-2504-19413}, and MIRIX for multi-dimensional organization~\citep{DBLP:journals/corr/abs-2507-07957}, and SCM for self-controlled memory management~\citep{10.1007/978-981-95-4158-4_12}. Recent work like MMS~\citep{DBLP:journals/corr/abs-2508-15294} and RMM~\citep{le2025rmm} further incorporates cognitive principles into memory design.
However, these prompting-based frameworks rely on expensive closed-source models and operate without feedback on downstream utility.

\subsection{Training-Based Approaches for Memory Construction}
Training-based methods can be categorized by memory form. Latent memory approaches encode information into hidden states: MEM1 consolidates memory through internal state mechanisms via RL~\citep{DBLP:journals/corr/abs-2506-15841}, while MemGen generates latent memory tokens within the reasoning stream~\citep{DBLP:journals/corr/abs-2509-24704}, 
and LongMem~\citep{DBLP:conf/nips/Wang0CLYGW23} uses a decoupled architecture with a frozen 
backbone as memory encoder and an adaptive side-network as memory retriever. 
Though efficient, these implicit representations sacrifice interpretability 
and fine-grained controllability.

Explicit memory approaches train models to manage structured external stores. Memory-R1 employs RL with sparse trajectory-level rewards~\citep{DBLP:journals/corr/abs-2508-19828}, but lacks learning signals for dense memory operations. Mem-$\alpha$ trains multi-dimensional memory construction, generalizing from 30k to 400k+ tokens~\citep{DBLP:journals/corr/abs-2509-25911}, yet applies a global reward across all memory operations regardless of their downstream impact.
While RLVR with GRPO~\citep{DBLP:journals/corr/abs-2402-03300} has become widely adopted~\citep{zhang2025100daysdeepseekr1survey}, sparse rewards remain insufficient for long-term dialogues, motivating our ADRPO.

\section{Methodology}

\subsection{Problem Formulation}

We address the task of long-term dialogue question answering. Given a sequence of conversation sessions $\mathcal{S} = \{s_1, s_2, \ldots, s_n\}$ with associated timestamps $\{t_1, t_2, \ldots, t_n\}$, and a question $q$ posed at time $t_q$ where $t_q > t_n$, the goal is to generate an accurate answer based on information distributed across the entire conversation history. Since concatenating all sessions typically exceeds context limits, we introduce an external memory bank $\mathcal{M}$ that compresses and organizes historical information for selective retrieval at inference time.
\subsection{Multi-Dimensional Memory Architecture}
\label{sec:architecture}

\begin{figure*}
    \centering
    \includegraphics[width=\textwidth, page=2]{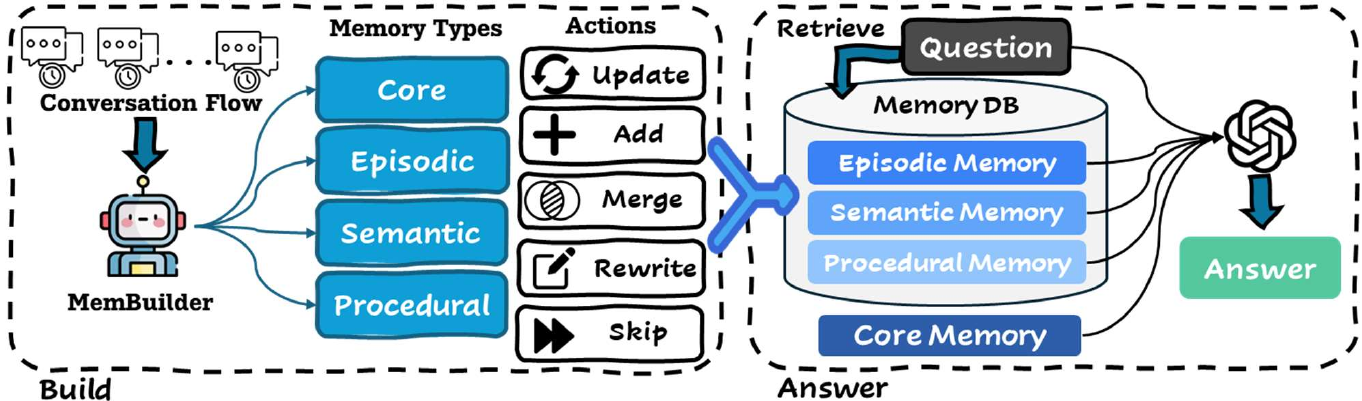}
    \caption{Multi-Dimensional Memory Architecture. Four memory types (Core, Episodic, Semantic, Procedural) are constructed during the Build Phase and selectively retrieved during the Answer Phase.}
    \label{fig:frame}
\end{figure*}

To effectively manage long-term dialogue information, we design a multi-dimensional memory system that decomposes conversations into four specialized memory types, each handled by a role-specific prompt to the same LLM (Figure~\ref{fig:frame}).

\paragraph{Memory Structure.} Our memory bank $\mathcal{M}$ consists of four components:
\begin{itemize}
    \item \textbf{Core Memory} $\mathcal{M}^{\textrm{core}}$: A fixed-size block storing persistent user profile information including identity, preferences, and key relationships. This memory is always included in the context during question answering. Detailed prompt templates are provided in Appendix~\ref{app:prompts}.
    
    \item \textbf{Episodic Memory} $\mathcal{M}^{\textrm{epi}}$: Time-stamped event records capturing what happened and when. Each entry follows the format ``\texttt{YYYY-MM-DD: Event summary | Details}'', enabling temporal reasoning.
    
    \item \textbf{Semantic Memory} $\mathcal{M}^{\textrm{sem}}$: Factual knowledge about entities in the user's life, such as people, places, and user-specific concepts. Common knowledge is explicitly excluded to avoid redundancy.
    
    \item \textbf{Procedural Memory} $\mathcal{M}^{\textrm{proc}}$: Step-by-step processes, routines, and workflows mentioned in conversations, such as the user's morning routine or problem-solving approach.

\end{itemize}

Given a new conversation session, all four memory types are processed simultaneously, each extracting memories according to its specialized perspective. Core Memory is maintained as a fixed block with automatic compression when capacity is exceeded. The other three memory types are stored in a vector database and retrieved via semantic similarity during question answering.

\paragraph{Memory Operations.} Since Core Memory operates on a single text block while the other three manage independent entries (Section \ref{sec:architecture}), their action spaces differ accordingly:
\begin{align}
\mathcal{A}^{\textrm{core}} &= \{\textsc{Append}, \textsc{Replace}, \textsc{Rewrite}\} \\
\mathcal{A}^{\textrm{epi}} &= \{\textsc{Add}, \textsc{Update}, \textsc{Merge}\} \\
\mathcal{A}^{\textrm{sem}} &= \{\textsc{Add}, \textsc{Update}, \textsc{Skip}\} \\
\mathcal{A}^{\textrm{proc}} &= \{\textsc{Add}, \textsc{Update}\}
\end{align}
At session $\tau$, given state $\mathit{State}_\tau = (\mathcal{M}_{\tau-1}, s_\tau)$, the LLM selects an action $a \in \mathcal{A}^{(m)}$ for each memory type and generates the corresponding memory content.

For Core Memory, \textsc{Append} adds new information to the block, \textsc{Replace} updates specific fragments, and \textsc{Rewrite} reorganizes the entire block. For the other three types, \textsc{Add} creates a new entry, and \textsc{Skip} bypasses common knowledge already captured in the model's parameters.

Unlike prior memory systems that delete old entries and replace them with new ones \citep{DBLP:journals/corr/abs-2504-19413,DBLP:journals/corr/abs-2507-07957}, we introduce two operations that preserve temporal history. The \textbf{\textsc{Update}} operation creates a new entry with a fresh timestamp that explicitly references the previous entry, rather than overwriting it, enabling the model to trace how information evolved. The \textbf{\textsc{Merge}} operation synthesizes multiple related events into a conclusion spanning a time range while preserving references to the original events as evidence, pre-computing complex temporal reasoning to reduce the burden during question answering.

This architecture transforms unstructured dialogues into organized, queryable memory. The remaining challenge is how to train the LLM to construct memory that maximizes downstream QA performance.

\subsection{Supervised Fine-Tuning}
While multi-dimensional memory architectures can offer finer-grained control \citep{DBLP:journals/corr/abs-2507-07957}, lightweight models, such as Qwen3-4B employed in our framework, often struggle with direct multi-dimensional memory construction, frequently producing invalid actions---such as malformed JSON structures, missing required fields (e.g., episodic entries without timestamps), or misuse of undefined operation types---that impede effective RL exploration. To address this cold-start problem, we collect expert trajectories $\{(\mathit{State}_t, a_t)\}_{t=1}^{n}$ using Claude 4.5 Sonnet. This phase stabilizes the model's output format, providing a viable baseline for subsequent training. However, SFT, as a form of imitation learning, trains the model to replicate expert memory operations but does not directly optimize for downstream QA accuracy; we further employ RL to maximize the model's ultimate utility in QA contexts.

\begin{figure*}
    \centering
    \includegraphics[width=\textwidth, page=3]{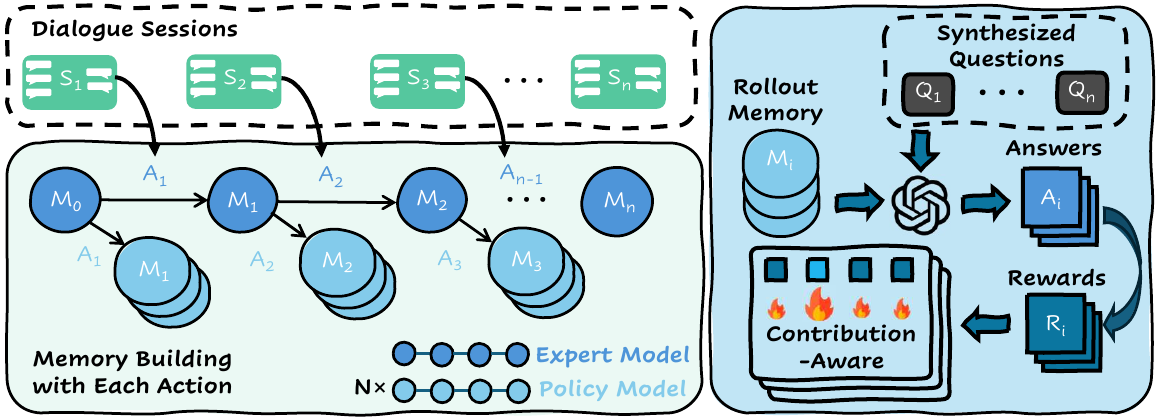}
    \caption{ADRPO training pipeline. Each session's memory rollouts are evaluated via synthetic QA, with gradients weighted by each memory component's downstream contribution.}
    \label{fig:train}
\end{figure*}

\subsection{Attributed Dense Rewards Policy Optimization (ADRPO)}
While SFT enables valid action generation, the resulting policy lacks optimization for downstream QA utility. We introduce a reinforcement learning algorithm that addresses two key challenges: \textbf{sparse trajectory-level rewards} in long-term dialogues, and \textbf{multi-dimensional memory attribution} among memory components with varying downstream impacts. Figure~\ref{fig:train} illustrates the ADRPO training pipeline.

% While SFT enables valid action generation, the resulting policy lacks optimization toward downstream QA utility. We introduce a reinforcement learning algorithm that addresses two key challenges (Figure~\ref{fig:train}): \textbf{temporal credit assignment} across extended dialogue histories and \textbf{multi-agent attribution} among collaborating agents with varying downstream impacts.

\subsubsection{Dense Session-Level Rewards via Synthetic Session-level QA}
\label{sec:dense_reward}

Prior RL approaches for memory construction~\citep{DBLP:journals/corr/abs-2508-19828,DBLP:journals/corr/abs-2509-25911} assign a single reward at the trajectory's end based on the final QA result. For dialogues spanning dozens of sessions, this provides no learning signal for dense memory operations.

We address this through synthetic session-level QA that evaluates memory quality at each step. Before the RL training, for each session $\tau$, we retrieve the top-$k$ memories from $\mathcal{M}_{\tau-1}$ most similar to the session $s_\tau$, and let an expert model with $(s_\tau, \mathcal{M}_{\textrm{retrieved}})$ generate $J$ question-answer pairs $\{(q_j, \mathit{ans}_j)\}_{j=1}^{J}$ targeting information in $s_\tau$ or its connections to retrieved memory.

During the RL training at $s_\tau$, we sample $N$ rollouts. Each rollout $i$ produces memory operations for all four memory types, yielding a candidate memory bank $\mathcal{M}_{\tau}^{(i)}$. A capable model answers pre-generated questions by retrieving from $\mathcal{M}_{\tau}^{(i)}$, and an LLM judge assesses correctness against ground-truth $\mathit{ans}_j$. The task reward measures memory construction quality as the average QA accuracy:
\begin{equation}
r^{\text{task}} = \frac{1}{J} \sum_{j=1}^{J} \mathbbm{1}[\text{correct}(q_j)]
\end{equation}
The final reward incorporates two regularization terms:
\begin{equation}
\label{eq:reward}
r = \mathbbm{1}[\text{valid}] \cdot r^{\text{task}} \cdot (1 - \lambda \cdot \ell)
\end{equation}

\paragraph{Format Validity.} The indicator $\mathbbm{1}[\text{valid}]$ acts as a gate: outputs with malformed JSON structure, missing required fields, or undefined actions receive zero reward regardless of content quality.

\paragraph{Length Penalty.} The term $\ell \in [0,1]$, weighted by $\lambda$, regularizes the amount of memory content stored. Let $|\mathcal{M}^{(m)}_{\text{new}}|$ and $|\hat{\mathcal{M}}^{(m)}|$ denote the token counts of memories stored by the policy and expert for memory type $m$, respectively. For core Memory, let $\Delta_{\text{core}} = |\mathcal{M}^{(\text{core})}_{\tau}| - |\mathcal{M}^{(\text{core})}_{\tau-1}|$ be the token increment after the operation:
\begin{equation}
\ell^{(\text{core})} = 
\begin{cases}
0 & \textrm{if } \Delta_{\text{core}} \leq \theta_{\min} \\[3pt]
\frac{\Delta_{\text{core}} - \theta_{\min}}{\theta_{\max} - \theta_{\min}} & \textrm{if } \Delta_{\text{core}} \in (\theta_{\min}, \theta_{\max}) \\[3pt]
1 & \textrm{if } \Delta_{\text{core}} \geq \theta_{\max}
\end{cases}
\end{equation}
where $\theta_{\min}$ and $\theta_{\max}$ are the penalty-free and full-penalty thresholds. For the other memory types, let $\rho = |\mathcal{M}^{(m)}_{\text{new}}| / |\hat{\mathcal{M}}^{(m)}|$ and $\Delta = \big||\mathcal{M}^{(m)}_{\text{new}}| - |\hat{\mathcal{M}}^{(m)}|\big|$:
\begin{equation}
\ell^{(m)} = 
\begin{cases}
0 & \textrm{if } \Delta < \delta \textrm{ or } \rho \in [\gamma_l, \gamma_u] \\[3pt]
\frac{\rho - \gamma_u}{\gamma_{\max} - \gamma_u} & \textrm{if } \rho \in (\gamma_u, \gamma_{\max}] \\[3pt]
\frac{\gamma_l - \rho}{\gamma_l - \gamma_{\min}} & \textrm{if } \rho \in [\gamma_{\min}, \gamma_l) \\[3pt]
1 & \textrm{otherwise}
\end{cases}
\end{equation}
where $\delta$ is the minimum difference threshold, $[\gamma_l, \gamma_u]$ the tolerance range, and $\gamma_{\min}$, $\gamma_{\max}$ the full-penalty boundaries.

\subsubsection{Contribution-Aware Gradient Weighting}
\label{sec:ca_weight}

Within each rollout, actions operating on the four memory dimensions contribute to a shared memory bank and receive a global reward. 
However, the functional impact on downstream QA performance varies significantly across memory types; for instance, Episodic Memories may be frequently retrieved while Procedural Memories remain unused. To account for these discrepancies, we dynamically amplify gradient updates based on each component's downstream utility. %However, their contributions to downstream QA differ by the distinct utilization of four types of memories. For example, it is possible that episodic memories are frequently retrieved while procedural memories remain unused . We scale gradient updates based on each agent's attributable contribution.
Since vanilla GRPO computes a single advantage per rollout and applies it uniformly to all components, it cannot distinguish their individual contributions---a challenge analogous to credit assignment in cooperative multi-agent settings~\citep{foerster2018coma,li2022dae}. Drawing on prior work that scales gradient updates by downstream utility to resolve such attribution ambiguity~\citep{li2022dae,ren2018l2rw}, we dynamically amplify gradient updates based on each component's retrieval-based contribution.

During QA evaluation, we record retrieval counts $h^{(m)}$ for each memory type $m \in \{\textrm{epi}, \textrm{sem}, \textrm{proc}\}$ across all questions. The dominant contributing type is:
\begin{equation}
d = \arg\max_{m \in \{\textrm{epi}, \textrm{sem}, \textrm{proc}\}} h^{(m)}
\end{equation}
Gradient weights are assigned as:
\begin{equation}
w^{(m)} = 
\begin{cases}
\alpha & \textrm{if } m = d \\
1 & \textrm{otherwise}
\end{cases}
\end{equation}
where $\alpha > 1$ amplifies updates for the dominant contributor. Core Memory, which is always included in the context rather than retrieved, receives a fixed weight $w^{(\textrm{core})} = 1$. This mechanism ensures that memory types whose entries directly contributed to successful QA receive proportionally stronger reinforcement.

\subsubsection{Training Objective}

We formulate the ADRPO training objective by extending GRPO~\citep{DBLP:journals/corr/abs-2402-03300} with the attributed session-level reward. At each session $\tau$, we sample $N$ rollouts from the current policy. Each rollout $i$ invokes the model four times in parallel, producing memory operations $a_i = (a_i^{\text{core}}, a_i^{\text{epi}}, a_i^{\text{sem}}, a_i^{\text{proc}})$ for each memory type. All four memory types share the session-level reward $r_i$ defined in Eq.~\ref{eq:reward}, but receive differentiated gradient weights $w^{(m)}$ based on their retrieval-based attribution.

Advantages are computed via within-group normalization:
\begin{equation}
A_i = \frac{r_i - \mu}{\sigma + \epsilon} = \frac{\mathbbm{1}[\text{valid}_i] \cdot r^{\text{task}}_i \cdot (1 - \lambda \ell_i) - \mu}{\sigma + \epsilon}
\end{equation}
where $\mu$ and $\sigma$ are computed over the $N$ rollouts, and $\epsilon$ is a small constant for numerical stability. The training objective is:
\begin{multline}
\mathcal{J}(\theta) = \mathbb{E}\Bigg[\sum_{m} \frac{1}{|a_i^{(m)}|} \sum_{k=1}^{|a_i^{(m)}|} \min\Big( w^{(m)} \rho_{i,k}^{(m)} A_i, \\
w^{(m)} \cdot \text{clip}(\rho_{i,k}^{(m)}, 1\!-\!\epsilon, 1\!+\!\epsilon) A_i \Big)\Bigg] - \beta \cdot D_{\textrm{KL}}(\pi_\theta \| \pi_{\textrm{ref}})
\end{multline}
where $\rho_{i,k}^{(m)} = \pi_\theta / \pi_{\textrm{ref}}$ is the importance ratio for the $k$-th token of memory type $m$'s output. The contribution-aware weights $w^{(m)}$ scale both terms within the $\min$ operator, uniformly amplifying the objective for high-impact components while preserving the clipping mechanism for stability.

\section{Experiments}

\subsection{Experimental Setup}

\paragraph{Datasets.}
We evaluate MemBuilder on three long-term dialogue benchmarks:
\textbf{LongMemEval} \citep{DBLP:conf/iclr/WuWYZCY25}, which consists of user-assistant chat histories designed to evaluate the long-term memory capabilities of chat assistants;
\textbf{LoCoMo} \citep{DBLP:conf/acl/MaharanaLTBBF24}, which contains human-human conversations between fictional personas grounded on temporal event graphs, spanning up to 35 sessions;
and \textbf{PerLTQA} \citep{du-etal-2024-perltqa}, a dataset featuring 141 characters with rich personal profiles, social relationships, and life events.
We train exclusively on the LongMemEval subset. LoCoMo and PerLTQA serve as OOD test sets that differ in both dialogue format and domain. Detailed statistics and data construction procedures are provided in Appendix~\ref{app:dataset}.

\paragraph{Baselines.}
We compare against:
(1) \textbf{RAG-based}: Following common practice in RAG-based systems \citep{mastra2025rag}, we implement two retrieval granularities: RAG-Session chunks dialogues at session boundaries and retrieves complete sessions, while RAG-Utterance embeds individual utterances for fine-grained matching but returns the containing session to preserve conversational context;
(2) \textbf{Memory frameworks}: Mem0 \citep{DBLP:journals/corr/abs-2504-19413} and MIRIX \citep{DBLP:journals/corr/abs-2507-07957};
(3) \textbf{Training-based}: Memory-R1 \citep{DBLP:journals/corr/abs-2508-19828}, whose results are taken from the original paper due to unavailable code.

\paragraph{Implementation Details.}
We use Qwen3-4B-Instruct-2507 as our base model. GPT-4.1 serves as the LLM judge for evaluation. All retrieval uses text-embedding-3-small. To isolate memory construction quality, we fix the answer model to Claude 4.5 Sonnet unless otherwise specified. Training hyperparameters are provided in Appendix~\ref{app:training} and baseline details in Appendix~\ref{app:baseline}. Detailed configuration including embedding settings and action formats is in Appendix~\ref{app:config}. A detailed cost breakdown is provided in Appendix~\ref{app:cost}.

\subsection{Training Data Construction}
\label{sec:data_construction}

We use LongMemEval as our sole training source, sampling 50 dialogues for SFT and a separate 50 for RL. All other benchmarks serve as OOD test sets.

\paragraph{SFT Dataset.}
The 50 SFT dialogues comprise approximately 2,400 sessions. We collect expert trajectories using Claude 4.5 Sonnet by processing sessions sequentially: at step $k$, the model receives the new session $s_k$ along with the top-20 memories retrieved from the step $k{-}1$ memory bank, then generates memory operations for all four types. After each step, newly created memories are embedded and added to the vector database. Since each session produces four memory type outputs trained as separate examples, this yields 9,600 training samples with an average input length of 6,000 tokens and output length of 780 tokens.

\paragraph{RL Dataset.}
The 50 RL dialogues similarly comprise approximately 2,400 sessions. For each session $\tau$, we construct the policy model's input by retrieving the top-20 most relevant memories from the step $\tau{-}1$ vector database, concatenated with the new session $s_\tau$, mirroring the inference-time setup.

To enable dense session-level rewards, we generate 5 QA pairs per session (12,000 pairs total) using Claude 4.5 Opus. The generation model receives the new session along with the retrieved memories, allowing it to create questions that test both current session retention and connections to prior history. Questions span three types: single-session (testing current session content), multi-session (requiring cross-session aggregation), and temporal-reasoning (involving time-based inference).

\paragraph{Reward Computation.}
During GRPO training (8 rollouts per session, 5 epochs), each rollout $i$ at session $\tau$ produces candidate memory bank $M^{(i)}_\tau$. GPT-4.1-mini answers the pre-generated questions by retrieving from $M^{(i)}_\tau$, and an LLM judge (GPT-4.1-mini) evaluates correctness against ground-truth answers. The average accuracy across 5 questions yields the task reward $r_{\textrm{task}}$ (Eq.~\ref{eq:reward}). Within each rollout, all four memory types share this reward, differentiated through contribution-aware gradient weighting (Section~\ref{sec:ca_weight}). This produces 96,000 session-rollouts for policy optimization. A human evaluation on 100 sampled QA pairs confirms that 92\% of generated answers are factually correct (Appendix~\ref{app:qa_quality}).

\subsection{Main Results}
Table~\ref{tab:main} presents performance across three benchmarks. To isolate the effect of memory construction quality, we set the answer model to Claude 4.5 Sonnet across all methods and compare three categories of approaches: retrieval-based methods, prompting-based frameworks, and training-based methods. Note that Memory-R1 trains Llama-3.1-8B-Instruct as both the memory construction and answer model. For a fairer comparison, we also evaluate our method with Qwen3-4B as the answer model (Table~\ref{tab:answer_model}), which still achieves 81.12\% on LoCoMo, significantly outperforming Memory-R1 (62.74\%).

\begin{table}[ht]
\centering

\resizebox{\columnwidth}{!}{
\begin{tabular}{lccc}
\toprule
\textbf{Method} & \textbf{LoCoMo} & \textbf{LongMemEval} & \textbf{PerLTQA} \\
\midrule
\multicolumn{4}{l}{\textit{Retrieval-based Methods}} \\
RAG-Session & 70.35 & 66.75 & 79.21\\
RAG-Utterance & 74.87 & 69.00 & 77.23 \\
\midrule
\multicolumn{4}{l}{\textit{Prompting-based Memory Construction}} \\
Mem0 & 51.64 & 47.00 & 62.04 \\
MIRIX & 77.48 & 73.25 & 83.11 \\
Ours (GPT-4.1) & 79.91 & 78.50 & 91.74 \\
Ours (QwQ-32B) & 77.47 & 76.00 & 88.96 \\
Ours (Claude 4.5 Sonnet) & 82.61 & 85.50 & 92.59 \\
\midrule
\multicolumn{4}{l}{\textit{Training-based Memory Construction}} \\
Memory-R1$^\dagger$ & 62.74 & - & - \\
Ours (Qwen3-4B) & 68.07 & 56.00 & 76.85 \\
\quad + SFT & 81.74 & 84.25 & 91.67 \\
\quad + RL & 79.31 & 62.75 & 82.19 \\
\quad + SFT + RL & \textbf{84.23} & \textbf{85.75} & \textbf{93.14} \\
\bottomrule
\multicolumn{4}{l}{\footnotesize $^\dagger$ Results from the original paper with a different answer model.} \\
\end{tabular}
}
\caption{Performance comparison of different memory construction methods. ``Ours'' denotes our memory architecture with different memory construction models.}
\label{tab:main}
\end{table}

Our method achieves SOTA performance across all three benchmarks. On LoCoMo, our trained Qwen3-4B model achieves 84.23\%, surpassing the best prompting-based framework MIRIX (77.48\%) by 6.75 percentage points and outperforming Claude 4.5 Sonnet as the memory construction model (82.61\%). Similar trends are observed on LongMemEval (85.75\%) and PerLTQA (93.14\%), where our method also outperforms all baselines including Claude 4.5 Sonnet.
These results demonstrate that, although memory construction requires frequent model invocations across sessions, a well-trained 4B model can effectively replace expensive closed-source APIs.
Notably, our model is trained exclusively on LongMemEval, yet achieves strong performance on LoCoMo and PerLTQA, demonstrating robust generalization to OOD benchmarks with different dialogue structures and question types.

The training stage ablation reveals the complementary roles of SFT and RL. SFT alone improves the base model from 68.07\% to 81.74\% by enabling valid multi-dimensional outputs, while RL further boosts performance to 84.23\% by optimizing for downstream QA utility. Notably, RL without SFT (79.31\%) underperforms SFT alone, confirming that supervised fine-tuning is essential to address the cold-start problem before effective RL exploration can proceed.

\subsection{Ablation Studies}
We conduct ablation experiments to analyze the contribution of our key design choices. Implementation details are provided in Appendix~\ref{app:ablation}.

\subsubsection{Effect of Gradient Weighting}
\begin{figure}[ht]
\centering
\includegraphics[width=\linewidth]{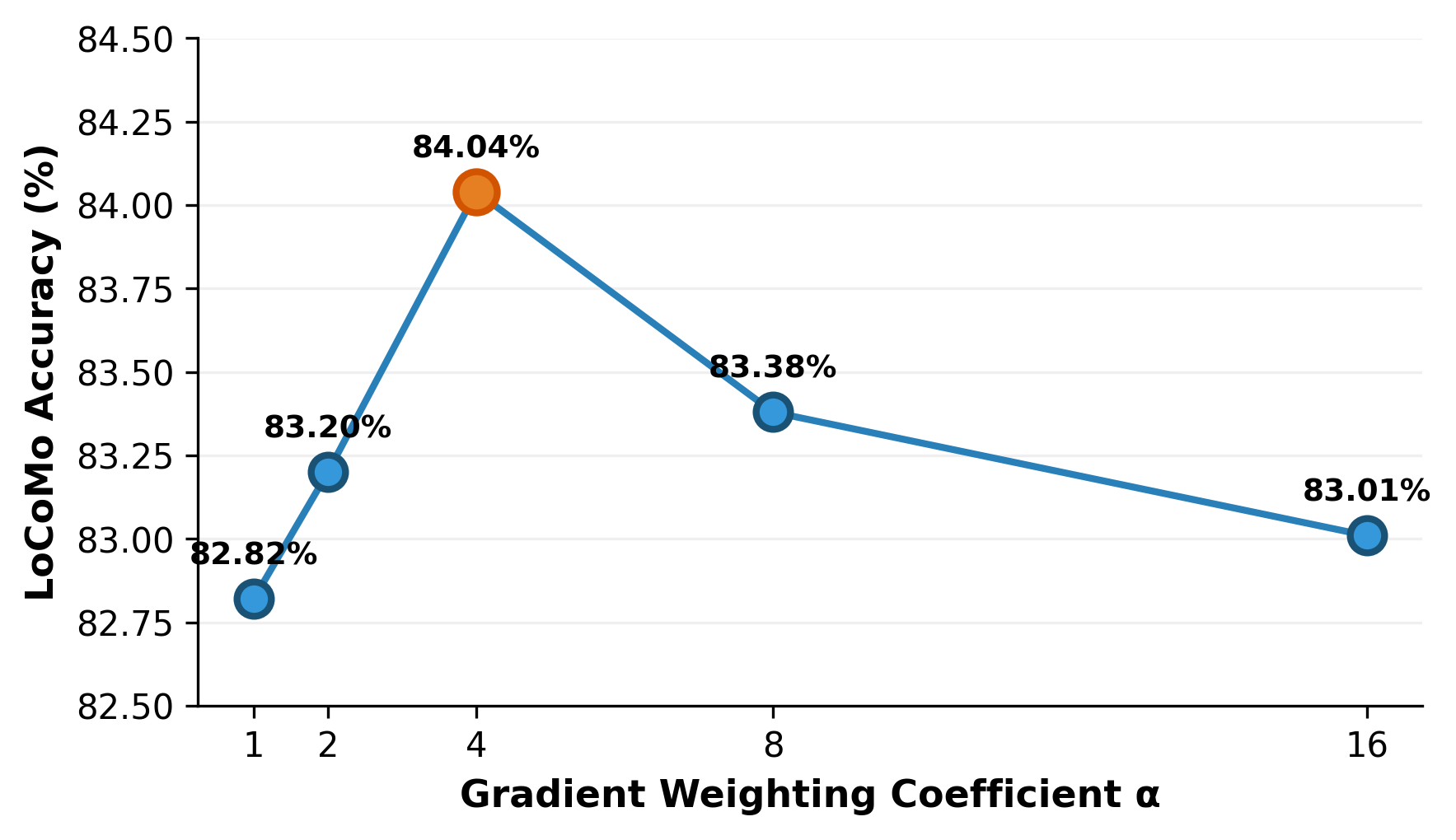}
\caption{Training curves with different gradient weighting coefficients $\alpha \in \{1, 2, 4, 8, 16\}$ on LoCoMo.}
\label{fig:alpha}
\end{figure}

To investigate the effect of contribution-aware gradient weighting (Section~\ref{sec:ca_weight}), we vary the weighting coefficient $\alpha$ that amplifies updates for the dominant contributing memory type. We conduct this ablation on a reduced training set for efficiency. As shown in Figure~\ref{fig:alpha}, performance improves as $\alpha$ increases from 1 (no weighting, 82.82\%) to 4 (84.04\%), confirming that attributing credit to high-contribution memory types enhances the final model performance. However, excessively large $\alpha$ values degrade performance due to gradient imbalance among memory types, with the optimal value at $\alpha=4$. A human evaluation further confirms that retrieval frequency is a reliable attribution proxy, with the dominant memory type aligning with the actual information source in 95\% of cases (Appendix~\ref{app:retrieval_validation}).

\subsubsection{Effect of Dense Rewards}
\begin{figure}[ht]
    \centering
    \includegraphics[width=1\linewidth]{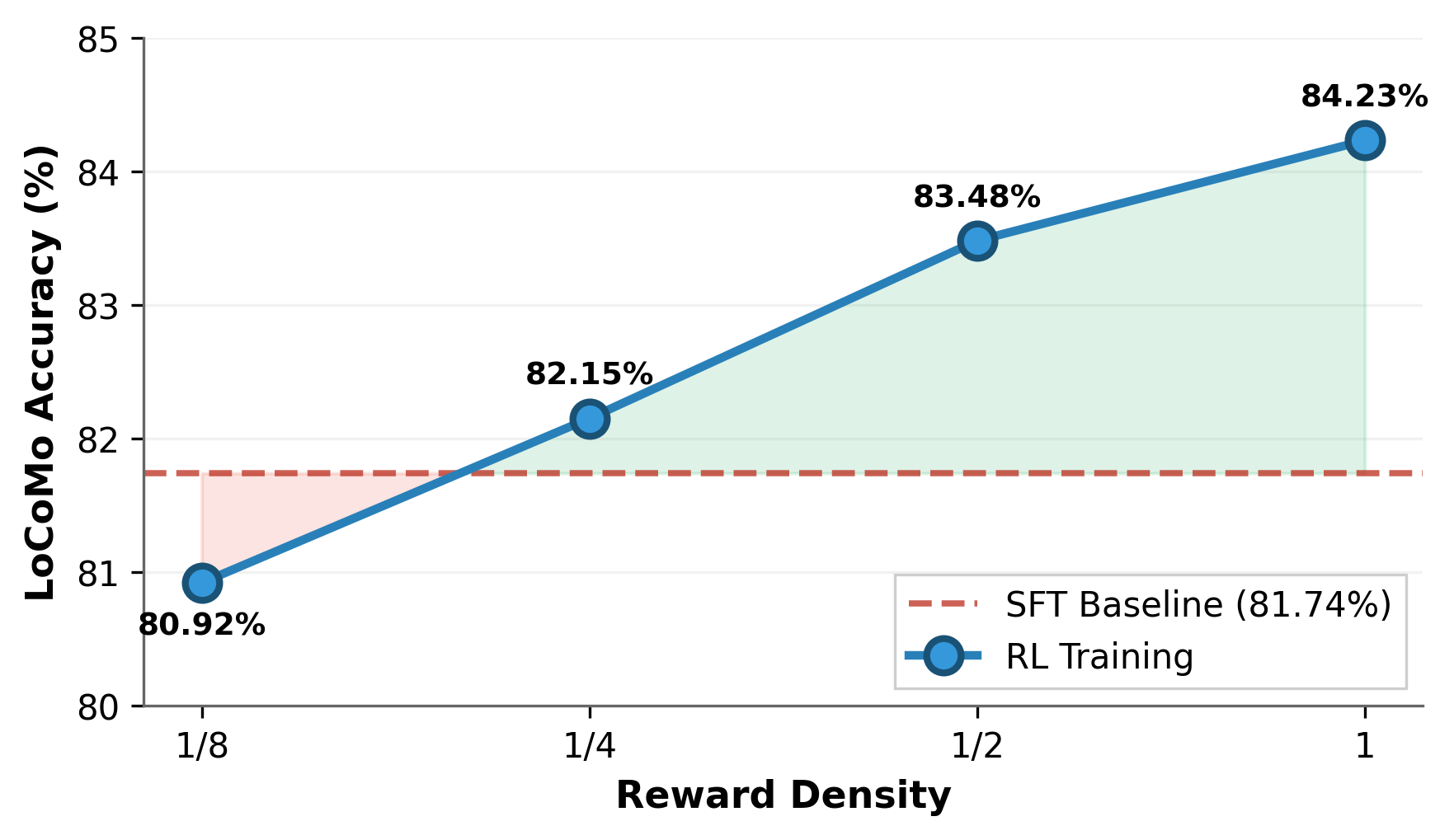}
    \caption{Effect of reward density on LoCoMo accuracy. The x-axis indicates the fraction of sessions receiving task rewards during training.}
    \label{fig:reward_density}
\end{figure}

To validate the importance of dense session-level rewards, we vary the reward density by providing task rewards to only a fraction of sessions during training. As shown in Figure~\ref{fig:reward_density}, under the same number of training epochs, model performance degrades consistently as the reward becomes sparser. When reward density drops to 1/8, performance falls below the SFT baseline (80.92\% vs 81.74\%).

These results reveal that sparse rewards not only slow convergence but can also be worse than using SFT alone, explaining why prior sparse-reward approaches achieve limited gains despite employing larger base models.

\subsubsection{Answer Model Generalization}

\begin{table}[h]
\centering
\resizebox{\columnwidth}{!}{
\begin{tabular}{lccc}
\toprule
\textbf{Answer Model} & \textbf{LoCoMo} & \textbf{LongMemEval} & \textbf{PerLTQA} \\
\midrule
Claude 4.5 Sonnet & 84.23 & 85.75 & 93.14 \\
GPT-4.1  & 81.51 & 82.50 & 91.83 \\
Qwen3-4B Base & 74.61 & 75.00 & 83.93 \\
Qwen3-4B Ours & 81.12 & 83.00 & 91.19 \\
\bottomrule
\end{tabular}
}
\caption{Performance with different answer models using memory constructed by our trained Qwen3-4B model. ``Qwen3-4B Base'' denotes the base model, while ``Qwen3-4B Ours'' denotes our trained model.}
\label{tab:answer_model}
\end{table}

To evaluate whether the constructed memory generalizes across different answer models, we fix the memory construction model to our trained Qwen3-4B and vary the answer model. Table~\ref{tab:answer_model} shows that our memory maintains high quality across answer models of varying capabilities.
We further investigate the feasibility of replacing closed-source answer models with open-source alternatives in Appendix~\ref{app:open_source}, where results show that Qwen3-30B-A3B achieves performance comparable to GPT-4.1-mini on LoCoMo when evaluated on the same high-quality memory.

% Interestingly, our trained Qwen3-4B as a construction model (81.12\%) outperforms the base model (74.61\%) and remains competitive with GPT-4.1, suggesting RL training creates implicit alignment between memory structure and the model's reasoning patterns.

Interestingly, when using our trained Qwen3-4B as an answer model, accuracy improves from 74.61\% to 81.12\% over the base model and remains competitive with GPT-4.1, suggesting RL training creates implicit alignment between memory structure and the model's reasoning patterns.

\subsection{Further Analysis}

\subsubsection{Performance by Question Type}
\begin{table}[ht]
\centering

\resizebox{\linewidth}{!}{
\begin{tabular}{lccccc}
\toprule
\textbf{Method} & \textbf{SingleHop} & \textbf{MultiHop} & \textbf{OpenDomain} & \textbf{Temporal} & \textbf{Adversarial} \\
\midrule
RAG-Utterance & 68.75 & 51.35 & 84.29 & 69.23 & 85.11 \\
Memory-R1 & 59.83 & 53.01 & 68.78 & 51.55 & - \\
Ours & 82.27 & 77.88 & 84.66 & 71.71 & 90.58 \\
\bottomrule
\end{tabular}
}
\caption{Performance breakdown by question type on LoCoMo.}
\label{tab:question_type}
\end{table}
Table~\ref{tab:question_type} details performance across LoCoMo question categories. Our method achieves the largest gains on MultiHop questions (77.88\% vs 53.01\% for Memory-R1, +24.87pp) and Temporal questions (71.71\% vs 51.55\%, +20.16pp), both of which require synthesizing information across multiple sessions. On Adversarial questions, our method achieves 90.58\%, demonstrating robustness against misleading information.

\subsubsection{Action Distribution Analysis}
\label{act_dis}
\begin{figure}[ht]
    \centering
    \includegraphics[width=1\linewidth]{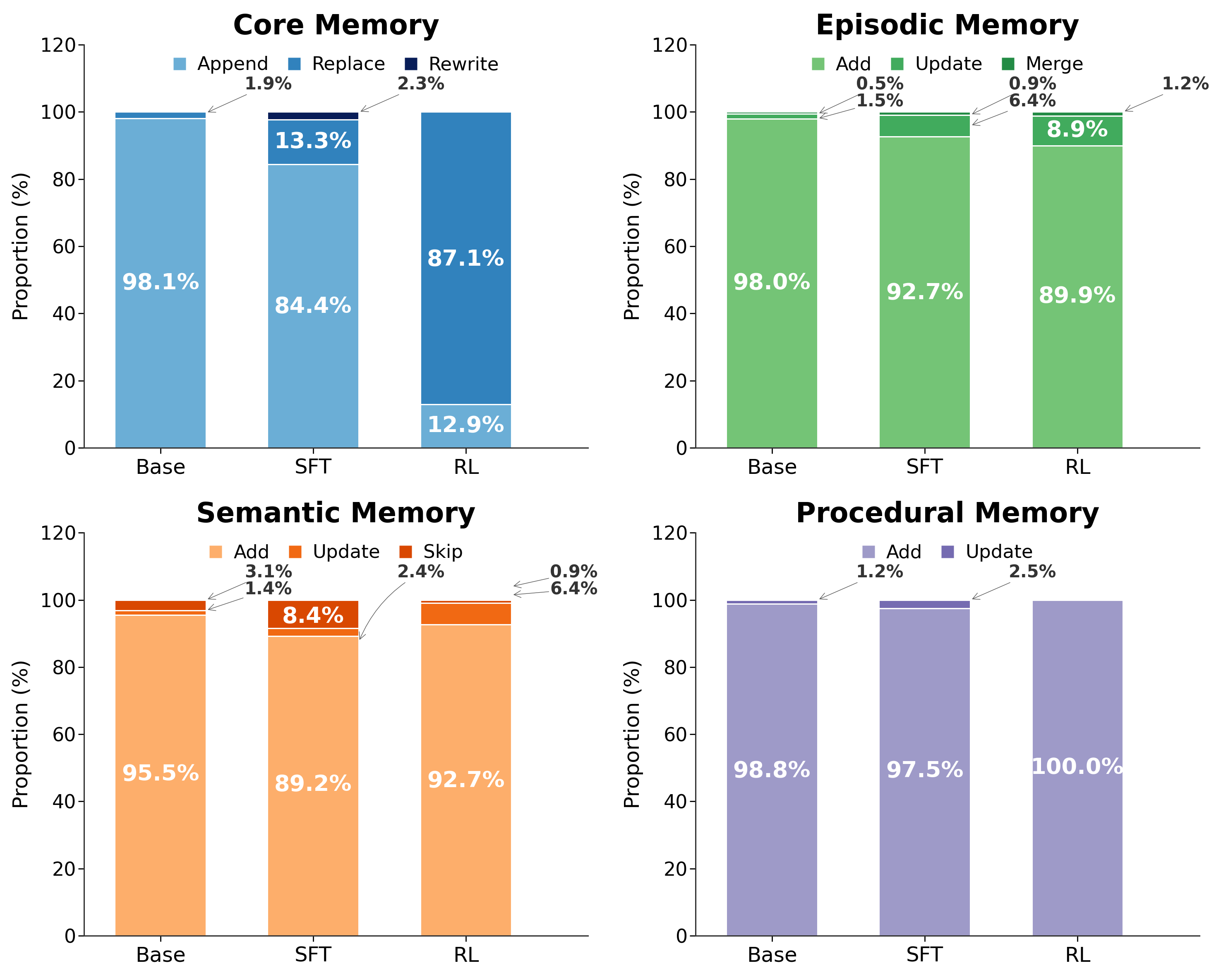}
    \caption{Action distribution across training stages (Base, SFT, RL) for four memory types. }
    \label{fig:action_dist}
\end{figure}

% Figure~\ref{fig:action_dist} visualizes how action distributions evolve across training stages.
% The most notable change is Core Memory's shift from \textsc{Append} (84.4\% $\rightarrow$ 12.9\%) to \textsc{Rewrite} (1.9\% $\rightarrow$ 87.1\%). Our analysis of the generated outputs demonstrates that the model learns to insert new details between existing sentences rather than appending at the end.
% The RL training also teaches the model to be more selective at generation time rather than relying on post-hoc filtering. For Semantic Memory, \textsc{Skip} operations decrease, indicating that the model directly outputs relevant facts rather than enumerating candidates and then excluding.
% We also observe divergent \textsc{Update} behavior: usage increases for Episodic and Semantic Memories but drops to zero for Procedural Memory, suggesting that evolving information (e.g., events and facts) benefits from explicit update chains while procedural knowledge is better maintained by adding discrete new entries.

Figure~\ref{fig:action_dist} visualizes how action distributions evolve across training stages. Concrete examples illustrating these behavioral changes are provided in Appendix~\ref{app:case_study}.

The most notable change is Core Memory's shift from \textsc{Append} (98.1\% $\rightarrow$ 12.9\%) to \textsc{Replace} (13.3\% $\rightarrow$ 87.1\%). Our analysis of the generated outputs demonstrates that the model learns to perform targeted updates to specific fields rather than appending at the end.
The RL training also teaches the model to be more selective at generation time rather than relying on post-hoc filtering. For Semantic Memory, \textsc{Skip} operations decrease, indicating that the model directly outputs relevant facts rather than enumerating candidates and then excluding.
We also observe divergent \textsc{Update} behavior: usage increases for Episodic and Semantic Memories but drops to zero for Procedural Memory, suggesting that evolving information (e.g., events and facts) benefits from explicit update chains while procedural knowledge is better maintained by adding discrete new entries.

\subsubsection{Training Dynamics}
\begin{figure}[ht]
\centering
\includegraphics[width=\columnwidth]{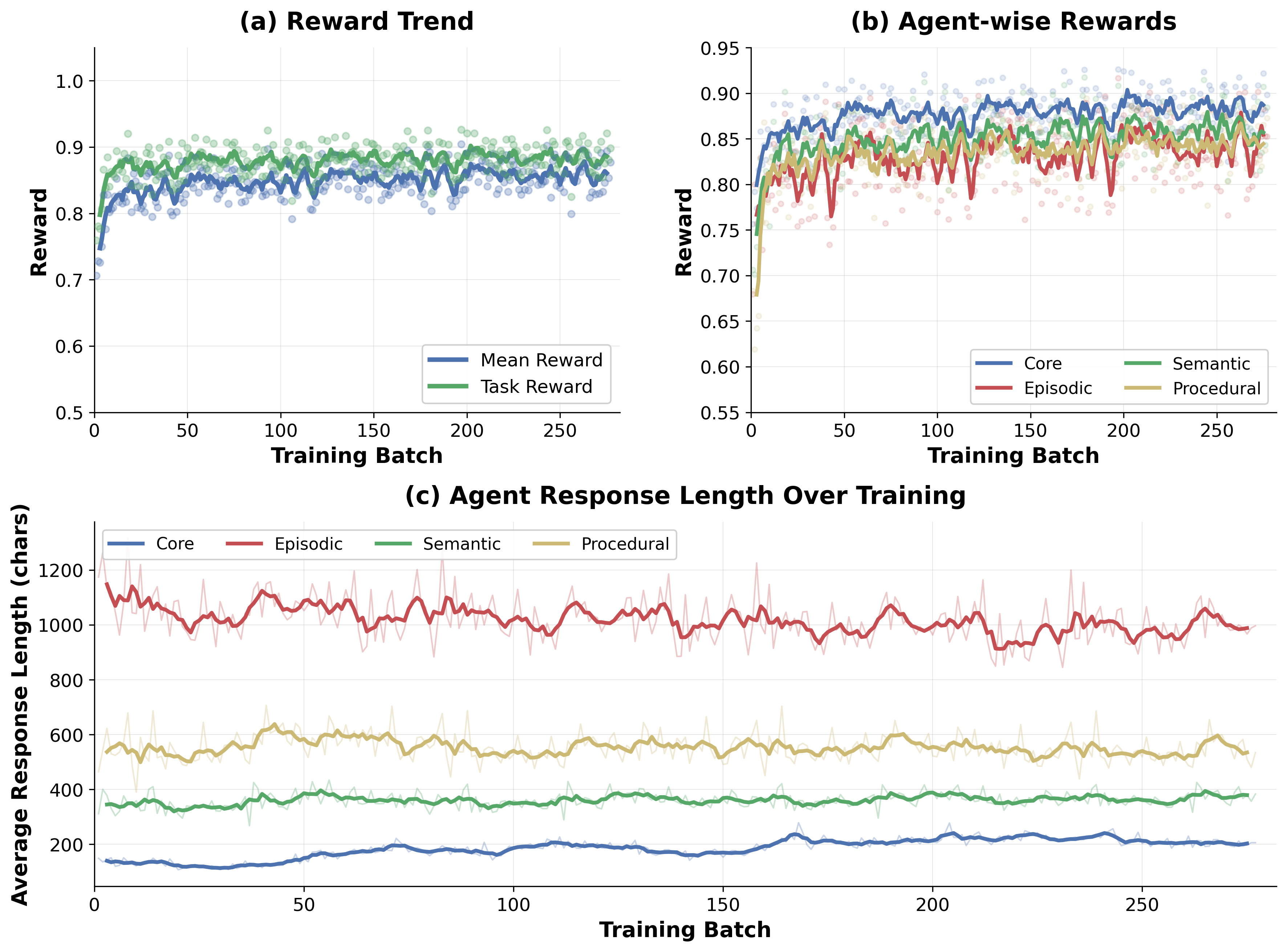}
\caption{Training dynamics: (a) overall reward trend, (b) rewards by memory type, and (c) response length by memory type. All metrics show stable improvement without reward hacking.}
\label{fig:training}
\end{figure}
Figure~\ref{fig:training} illustrates the training dynamics of our RL process. (a) Both task reward (QA accuracy) and mean reward (with length penalty) improve steadily, indicating effective learning. (b) All four memory types show consistent reward growth. (c) Response lengths remain stable throughout training, confirming that the length penalty prevents reward hacking through verbose outputs.

\section{Conclusion}
We presented MemBuilder, a reinforcement learning framework for multi-dimensional memory construction in long-term dialogues. By introducing ADRPO, Qwen3-4B achieves 84.23\% on LoCoMo, surpassing prompting-based frameworks using expensive closed-source models and generalizing effectively to OOD benchmarks. Our results demonstrate that memory construction can be handled by lightweight open-source models with appropriate training.

\section*{Limitations}

To isolate the impact of memory construction quality and ensure fair comparison across different methods, our evaluation relies on a fixed closed-source model (Claude 4.5 Sonnet) for question answering, which incurs API costs during evaluation. However, as shown in Table~\ref{tab:answer_model}, our constructed memory generalizes well across different answer models, suggesting that practitioners can substitute with capable open-source alternatives for cost-sensitive deployments. Furthermore, although we employ Claude 4.5 Opus for synthetic question generation, the generated QA pairs may still contain occasional inaccuracies or ambiguities. Despite this, our experimental results demonstrate that training with these synthetic questions substantially improves performance over sparse reward baselines.

\bibliography{custom}

@inproceedings{DBLP:conf/aaai/ZhongGGYW24,
  author       = {Wanjun Zhong and
                  Lianghong Guo and
                  Qiqi Gao and
                  He Ye and
                  Yanlin Wang},
  editor       = {Michael J. Wooldridge and
                  Jennifer G. Dy and
                  Sriraam Natarajan},
  title        = {MemoryBank: Enhancing Large Language Models with Long-Term Memory},
  booktitle    = {Thirty-Eighth {AAAI} Conference on Artificial Intelligence, {AAAI}
                  2024, Thirty-Sixth Conference on Innovative Applications of Artificial
                  Intelligence, {IAAI} 2024, Fourteenth Symposium on Educational Advances
                  in Artificial Intelligence, {EAAI} 2014, February 20-27, 2024, Vancouver,
                  Canada},
  pages        = {19724--19731},
  year         = {2024},
  url          = {https://doi.org/10.1609/aaai.v38i17.29946},
  doi          = {10.1609/AAAI.V38I17.29946},
  bibsource    = {dblp computer science bibliography, https://dblp.org}
}

@article{DBLP:journals/corr/abs-2310-08560,
  author       = {Charles Packer and
                  Vivian Fang and
                  Shishir G. Patil and
                  Kevin Lin and
                  Sarah Wooders and
                  Joseph E. Gonzalez},
  title        = {MemGPT: Towards LLMs as Operating Systems},
  journal      = {CoRR},
  volume       = {abs/2310.08560},
  year         = {2023},
  url          = {https://doi.org/10.48550/arXiv.2310.08560},
  doi          = {10.48550/ARXIV.2310.08560},
  eprinttype    = {arXiv},
  eprint       = {2310.08560},
  bibsource    = {dblp computer science bibliography, https://dblp.org}
}

@inproceedings{DBLP:conf/acl/MaharanaLTBBF24,
  author       = {Adyasha Maharana and
                  Dong{-}Ho Lee and
                  Sergey Tulyakov and
                  Mohit Bansal and
                  Francesco Barbieri and
                  Yuwei Fang},
  editor       = {Lun{-}Wei Ku and
                  Andre Martins and
                  Vivek Srikumar},
  title        = {Evaluating Very Long-Term Conversational Memory of {LLM} Agents},
  booktitle    = {Proceedings of the 62nd Annual Meeting of the Association for Computational
                  Linguistics (Volume 1: Long Papers), {ACL} 2024, Bangkok, Thailand,
                  August 11-16, 2024},
  pages        = {13851--13870},
  year         = {2024},
  url          = {https://doi.org/10.18653/v1/2024.acl-long.747},
  doi          = {10.18653/V1/2024.ACL-LONG.747},
  bibsource    = {dblp computer science bibliography, https://dblp.org}
}

@inproceedings{DBLP:conf/iclr/WuWYZCY25,
  author       = {Di Wu and
                  Hongwei Wang and
                  Wenhao Yu and
                  Yuwei Zhang and
                  Kai{-}Wei Chang and
                  Dong Yu},
  title        = {LongMemEval: Benchmarking Chat Assistants on Long-Term Interactive
                  Memory},
  booktitle    = {The Thirteenth International Conference on Learning Representations,
                  {ICLR} 2025, Singapore, April 24-28, 2025},
  year         = {2025},
  url          = {https://openreview.net/forum?id=pZiyCaVuti},
  bibsource    = {dblp computer science bibliography, https://dblp.org}
}

@inproceedings{du-etal-2024-perltqa,
    title = "{P}er{LTQA}: A Personal Long-Term Memory Dataset for Memory Classification, Retrieval, and Fusion in Question Answering",
    author = "Du, Yiming  and
      Wang, Hongru  and
      Zhao, Zhengyi  and
      Liang, Bin  and
      Wang, Baojun  and
      Zhong, Wanjun  and
      Wang, Zezhong  and
      Wong, Kam-Fai",
    editor = "Wong, Kam-Fai  and
      Zhang, Min  and
      Xu, Ruifeng  and
      Li, Jing  and
      Wei, Zhongyu  and
      Gui, Lin  and
      Liang, Bin  and
      Zhao, Runcong",
    booktitle = "Proceedings of the 10th SIGHAN Workshop on Chinese Language Processing (SIGHAN-10)",
    month = aug,
    year = "2024",
    address = "Bangkok, Thailand",
    url = "https://aclanthology.org/2024.sighan-1.18/",
    pages = "152--164"
}

@article{DBLP:journals/corr/abs-2309-12307,
  author       = {Yukang Chen and
                  Shengju Qian and
                  Haotian Tang and
                  Xin Lai and
                  Zhijian Liu and
                  Song Han and
                  Jiaya Jia},
  title        = {LongLoRA: Efficient Fine-tuning of Long-Context Large Language Models},
  journal      = {CoRR},
  volume       = {abs/2309.12307},
  year         = {2023},
  url          = {https://doi.org/10.48550/arXiv.2309.12307},
  doi          = {10.48550/ARXIV.2309.12307}
}

@inproceedings{peng2023yarn,
  author       = {Bowen Peng and
                  Jeffrey Quesnelle and
                  Honglu Fan and
                  Enrico Shippole},
  title        = {YaRN: Efficient Context Window Extension of Large Language Models},
  booktitle    = {The Twelfth International Conference on Learning Representations, {ICLR} 2024},
  year         = {2024},
  url          = {https://arxiv.org/abs/2309.00071}
}

@inproceedings{DBLP:conf/nips/TworkowskiSPWMM23,
  author       = {Szymon Tworkowski and
                  Konrad Staniszewski and
                  Mikolaj Pacek and
                  Yuhuai Wu and
                  Henryk Michalewski and
                  Piotr Milos},
  title        = {Focused Transformer: Contrastive Training for Context Scaling},
  booktitle    = {Advances in Neural Information Processing Systems 36: NeurIPS 2023},
  year         = {2023},
  url          = {http://papers.nips.cc/paper_files/paper/2023/hash/8511d06d5590f4bda24d42087802cc81-Abstract-Conference.html}
}

@inproceedings{bai2024longalign,
  author       = {Yushi Bai and
                  Xin Lv and
                  Jiajie Zhang and
                  Yuze He and
                  Ji Qi and
                  Lei Hou and
                  Jie Tang and
                  Yuxiao Dong and
                  Juanzi Li},
  title        = {LongAlign: A Recipe for Long Context Alignment of Large Language Models},
  booktitle    = {Findings of the Association for Computational Linguistics: EMNLP 2024},
  pages        = {1376--1395},
  year         = {2024},
  url          = {https://arxiv.org/abs/2401.18058}
}

@article{liu2024lost,
  author       = {Nelson F. Liu and
                  Kevin Lin and
                  John Hewitt and
                  Ashwin Paranjape and
                  Michele Bevilacqua and
                  Fabio Petroni and
                  Percy Liang},
  title        = {Lost in the Middle: How Language Models Use Long Contexts},
  journal      = {Transactions of the Association for Computational Linguistics},
  volume       = {12},
  pages        = {157--173},
  year         = {2024},
  url          = {https://arxiv.org/abs/2307.03172}
}

@incollection{tulving1972episodic,
  author       = {Endel Tulving},
  title        = {Episodic and Semantic Memory},
  booktitle    = {Organization of Memory},
  editor       = {Endel Tulving and Wayne Donaldson},
  year         = {1972},
  pages        = {381--403}
}

@article{sumers2024coala,
  author       = {Theodore R. Sumers and
                  Shunyu Yao and
                  Karthik Narasimhan and
                  Thomas L. Griffiths},
  title        = {Cognitive Architectures for Language Agents},
  journal      = {Transactions on Machine Learning Research},
  year         = {2024},
  url          = {https://arxiv.org/abs/2309.02427}
}

@book{laird2012soar,
  author       = {John E. Laird},
  title        = {The Soar Cognitive Architecture},
  year         = {2012},
  isbn         = {978-0-262-12296-5}
}

@article{DBLP:journals/corr/abs-2504-19413,
  author       = {Prateek Chhikara and
                  Dev Khant and
                  Saket Aryan and
                  Taranjeet Singh and
                  Deshraj Yadav},
  title        = {Mem0: Building Production-Ready {AI} Agents with Scalable Long-Term Memory},
  journal      = {CoRR},
  volume       = {abs/2504.19413},
  year         = {2025},
  url          = {https://doi.org/10.48550/arXiv.2504.19413},
  doi          = {10.48550/ARXIV.2504.19413}
}

@article{DBLP:journals/corr/abs-2507-07957,
  author       = {Yu Wang and
                  Xi Chen},
  title        = {{MIRIX:} Multi-Agent Memory System for LLM-Based Agents},
  journal      = {CoRR},
  volume       = {abs/2507.07957},
  year         = {2025},
  url          = {https://doi.org/10.48550/arXiv.2507.07957},
  doi          = {10.48550/ARXIV.2507.07957}
}

@article{DBLP:journals/corr/abs-2508-15294,
  author       = {Gaoke Zhang and
                  Bo Wang and
                  Yunlong Ma and
                  Dongming Zhao and
                  Zifei Yu},
  title        = {Multiple Memory Systems for Enhancing the Long-term Memory of Agent},
  journal      = {CoRR},
  volume       = {abs/2508.15294},
  year         = {2025},
  url          = {https://doi.org/10.48550/arXiv.2508.15294},
  doi          = {10.48550/ARXIV.2508.15294},
  eprinttype    = {arXiv},
  eprint       = {2508.15294},
  bibsource    = {dblp computer science bibliography, https://dblp.org}
}

@inproceedings{le2025rmm,
  author       = {Zhen Tan and
                  Tianshu Shen and
                  I-Hung Hsu and
                  Anbang Xu and
                  Longju Bai and
                  Yash Jain and
                  Chen-Yu Lee and
                  Hamid Palangi and
                  Tianlong Chen and
                  Long T. Le and
                  Rujun Han},
  title        = {In Prospect and Retrospect: Reflective Memory Management for 
                  Long-term Personalized Dialogue Agents},
  booktitle    = {Proceedings of the 63rd Annual Meeting of the Association for Computational Linguistics},
  pages        = {8416--8439},
  year         = {2025}
}

@article{DBLP:journals/corr/abs-2506-15841,
  author       = {Zijian Zhou and
                  Ao Qu and
                  Zhaoxuan Wu and
                  Sunghwan Kim and
                  Alok Prakash and
                  Daniela Rus and
                  Jinhua Zhao and
                  Bryan Kian Hsiang Low and
                  Paul Pu Liang},
  title        = {{MEM1:} Learning to Synergize Memory and Reasoning for Efficient Long-Horizon Agents},
  journal      = {CoRR},
  volume       = {abs/2506.15841},
  year         = {2025},
  url          = {https://doi.org/10.48550/arXiv.2506.15841},
  doi          = {10.48550/ARXIV.2506.15841}
}

@article{DBLP:journals/corr/abs-2509-24704,
  author       = {Guibin Zhang and
                  Muxin Fu and
                  Shuicheng Yan},
  title        = {MemGen: Weaving Generative Latent Memory for Self-Evolving Agents},
  journal      = {CoRR},
  volume       = {abs/2509.24704},
  year         = {2025},
  url          = {https://doi.org/10.48550/arXiv.2509.24704},
  doi          = {10.48550/ARXIV.2509.24704},
  eprinttype    = {arXiv},
  eprint       = {2509.24704},
  bibsource    = {dblp computer science bibliography, https://dblp.org}
}

@article{DBLP:journals/corr/abs-2508-19828,
  author       = {Sikuan Yan and
                  Xiufeng Yang and
                  Zuchao Huang and
                  Ercong Nie and
                  Zifeng Ding and
                  Zonggen Li and
                  Xiaowen Ma and
                  Hinrich Sch{\"{u}}tze and
                  Volker Tresp and
                  Yunpu Ma},
  title        = {Memory-R1: Enhancing Large Language Model Agents to Manage and Utilize
                  Memories via Reinforcement Learning},
  journal      = {CoRR},
  volume       = {abs/2508.19828},
  year         = {2025},
  url          = {https://doi.org/10.48550/arXiv.2508.19828},
  doi          = {10.48550/ARXIV.2508.19828},
  eprinttype    = {arXiv},
  eprint       = {2508.19828},
  bibsource    = {dblp computer science bibliography, https://dblp.org}
}

@article{DBLP:journals/corr/abs-2509-25911,
  author       = {Yu Wang and
                  Ryuichi Takanobu and
                  Zhiqi Liang and
                  Yuzhen Mao and
                  Yuanzhe Hu and
                  Julian J. McAuley and
                  Xiaojian Wu},
  title        = {Mem-{\(\alpha\)}: Learning Memory Construction via Reinforcement Learning},
  journal      = {CoRR},
  volume       = {abs/2509.25911},
  year         = {2025},
  url          = {https://doi.org/10.48550/arXiv.2509.25911},
  doi          = {10.48550/ARXIV.2509.25911},
  eprinttype    = {arXiv},
  eprint       = {2509.25911},
  bibsource    = {dblp computer science bibliography, https://dblp.org}
}

@article{DBLP:journals/corr/abs-2402-03300,
  author       = {Zhihong Shao and
                  Peiyi Wang and
                  Qihao Zhu and
                  Runxin Xu and
                  Junxiao Song and
                  Mingchuan Zhang and
                  Y. K. Li and
                  Y. Wu and
                  Daya Guo},
  title        = {DeepSeekMath: Pushing the Limits of Mathematical Reasoning in Open
                  Language Models},
  journal      = {CoRR},
  volume       = {abs/2402.03300},
  year         = {2024},
  url          = {https://doi.org/10.48550/arXiv.2402.03300},
  doi          = {10.48550/ARXIV.2402.03300},
  eprinttype    = {arXiv},
  eprint       = {2402.03300},
  bibsource    = {dblp computer science bibliography, https://dblp.org}
}

@misc{gao2024retrievalaugmentedgenerationlargelanguage,
      title={Retrieval-Augmented Generation for Large Language Models: A Survey}, 
      author={Yunfan Gao and Yun Xiong and Xinyu Gao and Kangxiang Jia and Jinliu Pan and Yuxi Bi and Yi Dai and Jiawei Sun and Meng Wang and Haofen Wang},
      year={2024},
      eprint={2312.10997},
      archivePrefix={arXiv},
      primaryClass={cs.CL},
      url={https://arxiv.org/abs/2312.10997}, 
}

@misc{zhang2025100daysdeepseekr1survey,
      title={100 Days After DeepSeek-R1: A Survey on Replication Studies and More Directions for Reasoning Language Models}, 
      author={Chong Zhang and Yue Deng and Xiang Lin and Bin Wang and Dianwen Ng and Hai Ye and Xingxuan Li and Yao Xiao and Zhanfeng Mo and Qi Zhang and Lidong Bing},
      year={2025},
      eprint={2505.00551},
      archivePrefix={arXiv},
      primaryClass={cs.CL},
      url={https://arxiv.org/abs/2505.00551}, 
}

@inproceedings{zheng-etal-2024-llamafactory,
    title = "{L}lama{F}actory: Unified Efficient Fine-Tuning of 100+ Language Models",
    author = "Zheng, Yaowei  and
      Zhang, Richong  and
      Zhang, Junhao  and
      Ye, Yanhan  and
      Luo, Zheyan",
    editor = "Cao, Yixin  and
      Feng, Yang  and
      Xiong, Deyi",
    booktitle = "Proceedings of the 62nd Annual Meeting of the Association for Computational Linguistics (Volume 3: System Demonstrations)",
    month = aug,
    year = "2024",
    address = "Bangkok, Thailand",
    url = "https://aclanthology.org/2024.acl-demos.38/",
    doi = "10.18653/v1/2024.acl-demos.38",
    pages = "400--410"
}

@inproceedings{DBLP:conf/eurosys/ShengZYWZZPL025,
  author       = {Guangming Sheng and
                  Chi Zhang and
                  Zilingfeng Ye and
                  Xibin Wu and
                  Wang Zhang and
                  Ru Zhang and
                  Yanghua Peng and
                  Haibin Lin and
                  Chuan Wu},
  title        = {HybridFlow: {A} Flexible and Efficient {RLHF} Framework},
  booktitle    = {Proceedings of the Twentieth European Conference on Computer Systems,
                  EuroSys 2025, Rotterdam, The Netherlands, 30 March 2025 - 3 April
                  2025},
  pages        = {1279--1297},
  year         = {2025},
  url          = {https://doi.org/10.1145/3689031.3696075},
  doi          = {10.1145/3689031.3696075},
  bibsource    = {dblp computer science bibliography, https://dblp.org}
}

@misc{mastra2025rag,
  author = {Mastra},
  title = {Yes, you can use RAG for agent memory},
  year = {2025},
  howpublished = {\url{https://mastra.ai/blog/use-rag-for-agent-memory}},
  note = {Accessed: 2025}
}

@InProceedings{10.1007/978-981-95-4158-4_12,
author="Wang, Bing
and Liang, Xinnian
and Yang, Jian
and Huang, Hui
and Wu, Zhenhe
and Wu, ShuangZhi
and Ma, Zejun
and Li, Zhoujun",
editor="Zhu, Feida
and Yu, Philip S.
and Nadamoto, Akiyo
and Lim, Ee-Peng
and Shim, Kyuseok
and Ding, Wei
and Zhang, Bingxue",
title="SCM: Enhancing Large Language Model with Self-Controlled Memory Framework",
booktitle="Database Systems for Advanced Applications",
year="2026",
address="Singapore",
pages="188--203",
isbn="978-981-95-4158-4"
}

@inproceedings{DBLP:conf/nips/Wang0CLYGW23,
  author       = {Weizhi Wang and
                  Li Dong and
                  Hao Cheng and
                  Xiaodong Liu and
                  Xifeng Yan and
                  Jianfeng Gao and
                  Furu Wei},
  editor       = {Alice Oh and
                  Tristan Naumann and
                  Amir Globerson and
                  Kate Saenko and
                  Moritz Hardt and
                  Sergey Levine},
  title        = {Augmenting Language Models with Long-Term Memory},
  booktitle    = {Advances in Neural Information Processing Systems 36: Annual Conference
                  on Neural Information Processing Systems 2023, NeurIPS 2023, New Orleans,
                  LA, USA, December 10 - 16, 2023},
  year         = {2023},
  url          = {http://papers.nips.cc/paper\_files/paper/2023/hash/ebd82705f44793b6f9ade5a669d0f0bf-Abstract-Conference.html},
  bibsource    = {dblp computer science bibliography, https://dblp.org}
}

@inproceedings{foerster2018coma,
  author    = {Jakob N. Foerster and
               Gregory Farquhar and
               Triantafyllos Afouras and
               Nantas Nardelli and
               Shimon Whiteson},
  title     = {Counterfactual Multi-Agent Policy Gradients},
  booktitle = {Proceedings of the Thirty-Second {AAAI} Conference on Artificial
               Intelligence ({AAAI-18})},
  pages     = {2974--2982},
  year      = {2018},
}

@inproceedings{li2022dae,
  author    = {Yueheng Li and
               Guangming Xie and
               Zongqing Lu},
  title     = {Difference Advantage Estimation for Multi-Agent Policy Gradients},
  booktitle = {Proceedings of the 39th International Conference on Machine Learning
               ({ICML} 2022)},
  series    = {Proceedings of Machine Learning Research},
  volume    = {162},
  pages     = {13066--13085},
  year      = {2022},
}

@inproceedings{ren2018l2rw,
  author    = {Mengye Ren and
               Wenyuan Zeng and
               Bin Yang and
               Raquel Urtasun},
  title     = {Learning to Reweight Examples for Robust Deep Learning},
  booktitle = {Proceedings of the 35th International Conference on Machine Learning
               ({ICML} 2018)},
  series    = {Proceedings of Machine Learning Research},
  volume    = {80},
  pages     = {4334--4343},
  year      = {2018},
}

\appendix

\section{Dataset Details}
\label{app:dataset}

\subsection{Benchmark Datasets}

We evaluate on three long-term dialogue benchmarks that differ in dialogue format, domain, and data organization.

\textbf{LongMemEval} consists of user-assistant chat histories designed to evaluate the long-term memory capabilities of chat assistants. The dataset contains 500 independent questions, each with its own dialogue context, averaging 40 sessions and approximately 115K tokens. Questions test five core memory abilities: information extraction, multi-session reasoning, temporal reasoning, knowledge updates, and abstention. This benchmark serves as both our training source and in-distribution evaluation set.

\textbf{LoCoMo} contains human-human conversations between fictional personas, grounded on temporal event graphs. The dataset includes 10 dialogues with an average of 27 sessions (range: 19--32) and 14K tokens per dialogue. Each dialogue is associated with multiple questions, totaling 1,986 questions across five types: SingleHop (282), MultiHop (321), OpenDomain (841), Temporal (96), and Adversarial (446). Unlike LongMemEval's one-question-per-context format, LoCoMo tests memory systems on shared dialogue contexts with diverse question types.

\textbf{PerLTQA} is a dataset featuring 141 fictional characters, including 30 protagonists with rich personal profiles, social relationships, and life events. Questions are designed around the 30 protagonists, totaling 8,593 questions across five types: factual, reasoning, other, yes/no, and temporal. This dataset requires integrating Episodic Memories with Semantic Memories about characters.

Both LoCoMo and PerLTQA serve as out-of-distribution test sets, differing from LongMemEval in dialogue format (human-human vs. user-assistant), data organization (shared context vs. independent context), and domain.

% \subsection{Training Data Construction}

% We use LongMemEval as our sole training source. From the 500 available dialogues, we sample 50 dialogues for SFT trajectory collection and a separate 50 dialogues for RL dataset construction. All other benchmarks (LoCoMo and PerLTQA) serve as out-of-distribution test sets to evaluate generalization.

% \paragraph{SFT Dataset.}
% The 50 SFT dialogues comprise approximately 2,400 sessions. Since each session requires memory operations from all four memory types (Core, Episodic, Semantic, Procedural) and we train each memory type's output as a separate example, this yields 9,600 training samples. Each sample pairs an input (retrieved memories + current session) with a single type of memory operations in JSON format. Based on logged data, the average input length is approximately 6,000 tokens and the average output length is approximately 780 tokens, resulting in a total of approximately 65M tokens for SFT training.

% \paragraph{RL Dataset.}
% The 50 RL dialogues comprise approximately 2,400 sessions. For each session, we generate 5 synthetic QA pairs for dense reward computation, resulting in 12,000 QA pairs in total. During GRPO training with 8 rollouts per session and 5 epochs, this produces 96,000 session-rollouts for policy optimization.

\section{Training Details}
\label{app:training}

\subsection{SFT Training}

We perform supervised fine-tuning using LlamaFactory~\citep{zheng-etal-2024-llamafactory}. The training data consists of expert trajectories where each example pairs a session input (retrieved memories + new session) with the expert model's memory operations for all four memory types. We use Qwen3-4B-Instruct-2507 as the base model with learning rate $5 \times 10^{-7}$, batch size 4, and train for 10 epochs.

\subsection{RL Training}

We implement ADRPO by extending the verl framework~\citep{DBLP:conf/eurosys/ShengZYWZZPL025} with contribution-aware gradient weighting and session-level reward computation. Starting from the SFT checkpoint, we train with a learning rate $1 \times 10^{-6}$ and a batch size of 128. The rollout number is set to 8, and the clipping parameter $\epsilon$ is set to 0.2. For contribution-aware gradient weighting, we use $\alpha = 4$ based on ablation results (Figure~\ref{fig:alpha}). The length penalty coefficient $\lambda$ is set to 0.8, with Core memory thresholds $\theta_{\min} = 150$ and $\theta_{\max} = 400$, and other memory parameters $\delta = 200$ and $[\gamma_l, \gamma_u] = [0.5, 1.3]$. Training runs for 5 epochs on 32 H20 GPUs (4 nodes), taking approximately 70 hours.

\section{Framework Configuration}
\label{app:config}

\subsection{Embedding and Retrieval}

We use text-embedding-3-small as the embedding model for all retrieval operations throughout the framework. For memory construction, we retrieve the top-20 most relevant old memories to provide context for the current session. For QA answering, we similarly retrieve the top-10 memories from the final memory bank. The RAG baselines (RAG-Session and RAG-Utterance) retrieve the top-5 chunks. For synthetic QA generation during RL dataset construction, we retrieve the top-20 memories to provide richer historical context.

\subsection{Memory Configuration}

\paragraph{Core memory} is maintained as a single text block with a maximum capacity of 5,000 characters. When the content exceeds this limit after an Append or Replace operation, the same policy model is prompted to compress the content while preserving essential information, including user identity, key relationships, personality traits, important preferences, and long-term goals. The compression removes redundant descriptions, minor details, and verbose explanations. If the first compression attempt still exceeds the limit, a second, more aggressive compression pass is performed.

\paragraph{Episodic, Semantic, and Procedural Memories} are stored as individual entries in a vector database with no explicit size limit. Each entry is embedded independently for retrieval.

\subsection{Agent Action Format}

We refer to the role-specific prompt for each memory type as an \textit{agent}. For each memory type, the model receives the current session along with relevant retrieved memories and outputs a JSON object specifying the action type and content. Below, we describe the action format for each agent.

\paragraph{Core memory Agent.}
The Core memory Agent manages persistent user information, including identity, preferences, personality traits, and key relationships. It outputs one of three operations:

\begin{itemize}
    \item \textbf{APPEND}: Add new information to the existing Core memory block (used when capacity $<$90\%).
    \item \textbf{REPLACE}: Update specific outdated or incorrect information by specifying old and new text.
    \item \textbf{REWRITE}: Reorganize and consolidate the entire block (used when capacity $>$90\% or when major updates are needed).
\end{itemize}

Example output:
\begin{lstlisting}
{"operation": "APPEND", 
 "content": "Works as a software engineer at Google, specializing in machine learning"}
\end{lstlisting}

\paragraph{Episodic Memory Agent.}
The Episodic Memory Agent manages time-ordered event memories. Each entry includes a timestamp, summary, and detailed description capturing who, what, when, where, and why. It outputs operations from:

\begin{itemize}
    \item \textbf{ADD}: Create a new event entry not currently in memory.
    \item \textbf{UPDATE}: Add a new related event that references previous events (old versions remain for history).
    \item \textbf{MERGE}: Combine multiple related events into a timeline with a timestamp range, drawing conclusions from patterns (old versions remain for history).
\end{itemize}

Example output:
\begin{lstlisting}
{"operations": [
  {"action": "ADD", 
   "memory": "2024-03-15: Started new job at startup | Details: First day at TechCorp as senior engineer, met team lead Sarah..."}
]}
\end{lstlisting}

\paragraph{Semantic Memory Agent.}
The Semantic Memory Agent manages conceptual knowledge about people, places, objects, and concepts in the user's life. It explicitly skips common knowledge already captured in the model's parameters. Operations include:

\begin{itemize}
    \item \textbf{ADD}: Create an entry for a new concept, person, or object.
    \item \textbf{UPDATE}: Add new information to an existing concept.
    \item \textbf{SKIP}: Bypass common knowledge or already fully captured information.
\end{itemize}

Example output:
\begin{lstlisting}
{"operations": [
  {"action": "ADD", 
   "memory": "Sarah (colleague) - Career: Team lead at TechCorp, 5 years experience in ML..."},
  {"action": "SKIP", 
   "reason": "Common knowledge about Python"}
]}
\end{lstlisting}

\paragraph{Procedural Memory Agent.}
The Procedural Memory Agent manages step-by-step processes, workflows, and instructions. Each entry includes a description, numbered steps, and optional context. Operations include:

\begin{itemize}
    \item \textbf{ADD}: Create a new procedure entry.
    \item \textbf{UPDATE}: Modify an existing procedure with new information.
\end{itemize}

Example output:
\begin{lstlisting}
{"operations": [
  {"action": "ADD", 
   "memory": "Morning workout routine | Steps: 1. 10min stretching 2. 30min jogging 3. 15min core exercises | Context: Daily routine before work"}
]}
\end{lstlisting}

\paragraph{Validity Criteria.}
An action is considered \textbf{valid} if: (1) the JSON structure is well-formed, (2) the action type is defined for that agent, (3) all required fields are present, and (4) for UPDATE/MERGE operations, referenced entries exist in the current memory bank. Invalid actions receive zero reward regardless of content quality.

\section{Baseline Implementation}
\label{app:baseline}

\subsection{RAG Baselines}

\textbf{RAG-Session} segments dialogues at session boundaries, treating each session as a retrieval unit. Given a question, we retrieve the top-5 most similar sessions using text-embedding-3-small and provide them as context to the answer model.
\textbf{RAG-Utterance} segments dialogues at the utterance level, treating each user-assistant turn pair as a retrieval unit. Given a question, we retrieve the top-5 most similar utterances using text-embedding-3-small and provide them as context to the answer model.

\subsection{Memory Frameworks}

For fair comparison, we evaluate all memory frameworks using Claude 4.5 Sonnet as both the memory construction model and answer model, except for ablations in Table~\ref{tab:answer_model} which vary the answer model.

\textbf{Mem0} is configured with its default settings for memory extraction and organization.

\textbf{MIRIX} is configured with its default multi-dimensional memory structure.

Both frameworks construct memories by processing dialogues sequentially, then answer questions by retrieving from the constructed memory bank.

\subsection{Training-based Methods}

\textbf{Memory-R1} results are taken from the original paper. Note that Memory-R1 uses Llama-3.1-8B-Instruct as both the memory construction and answer model, which differs from our evaluation setup. For reference, our method with Qwen3-4B as both the construction and answer model achieves 82.00\% on LoCoMo (Table~\ref{tab:answer_model}), substantially outperforming Memory-R1's reported 62.74\%.

\section{Ablation Experiment Details}
\label{app:ablation}

\subsection{Gradient Weighting Ablation}

To efficiently explore the effect of contribution-aware gradient weighting, we conduct this ablation on a reduced training set consisting of 10 dialogues sampled from the 50 RL training dialogues. We vary $\alpha \in \{1, 2, 4, 8, 16\}$ while keeping all other hyperparameters fixed. The setting $\alpha = 1$ corresponds to uniform weighting without contribution-aware scaling. Results are shown in Figure~\ref{fig:alpha}.

\subsection{Reward Density Ablation}

We conduct this ablation on the full training set of 50 dialogues. To validate the importance of dense session-level rewards, we vary the reward density by randomly skipping task reward computation for a fraction of sessions. Specifically, at density $1/d$, each session independently has a probability $1/d$ of receiving a task reward. Sessions without task rewards still receive format validity and length penalty signals, but no QA-based feedback. All configurations are trained for the same number of epochs to ensure fair comparison. Results are shown in Figure~\ref{fig:reward_density}.

\section{Cost Analysis}
\label{app:cost}

Our training pipeline incurs API costs at two stages: (1) data preparation (one-time) and (2) RL training (per-run).

\subsection{Data Preparation Costs (One-Time)}

\paragraph{Expert Trajectory Generation.}
We use Claude 4.5 Sonnet to generate memory management demonstrations for 50 LongMemEval conversations (2,400 sessions). Each session invokes 4 memory agent calls (Core, Episodic, Semantic, Procedural), resulting in 9,600 API calls. Based on actual logged data from 952 agent calls, the average token consumption is 6,011 input tokens (agent prompt + existing memories + current session) and 784 output tokens (JSON memory operations). Total: $\sim$58M input + $\sim$8M output tokens. Cost: \textbf{\$294}.

\paragraph{Synthetic Question Generation.}
For all 50 conversations (2,400 sessions), we generate 5 QA pairs per session using Claude 4.5 Opus. Based on analysis of generated QA files, each call uses $\sim$3,500 input tokens (QA prompt template + memory state + current session) and produces $\sim$460 output tokens (5 questions with answers in JSON). Total: $\sim$8.5M input + $\sim$1.1M output tokens. Cost: \textbf{\$70}.

\subsection{RL Training Costs (Per-Run)}

During GRPO training (2,400 sessions $\times$ 8 rollouts $\times$ 5 epochs = 96,000 session-rollouts), reward computation evaluates 5 QA questions per rollout, yielding 480,000 QA evaluations. Each evaluation involves two API calls: answer generation and LLM judging, both using GPT-4.1-mini. Answer generation consumes approximately 960M input and 48M output tokens (\$461), while LLM judging requires approximately 241M input and 15M output tokens (\$120). The per-run total is \textbf{\$581}.

\section{Case Study}
\label{app:case_study}

To illustrate the behavioral changes discussed in Section~\ref{act_dis}, we present representative examples comparing model outputs across training stages (Base, SFT, RL) on identical inputs.

\subsection{Core Memory: Precision Evolution}

\paragraph{Scenario.} The user discusses updates to their vegetarian lifestyle. The current Core Memory contains:

\begin{lstlisting}
Dietary preference: Exploring plant-based options
Family: Has children
Hobbies: Cooking, reading
\end{lstlisting}

\paragraph{User Message.} ``I've been vegetarian for about 6 months now. My kids love when I make vegan pancakes on Sunday mornings. We tried oat milk last week and it's now my favorite plant-based milk!''

\paragraph{Base Model.} Produces a fragmented append without structure:
\begin{lstlisting}
{"operation": "APPEND", "content": "vegetarian 6 months. oat milk favorite. vegan pancakes."}
\end{lstlisting}

\paragraph{SFT Model.} Rewrites the entire memory block:
\begin{lstlisting}
{"operation": "REWRITE", "content": "Dietary preference: Vegetarian for 6 months, exploring plant-based options; kids love vegan pancakes on Sunday mornings; tried oat milk last week and now it's favorite plant-based milk\nFamily: Has children\nHobbies: Cooking, reading"}
\end{lstlisting}

\paragraph{RL Model.} Performs a targeted replacement of only the relevant field:
\begin{lstlisting}
{"operation": "REPLACE", "old_text": "Dietary preference: Exploring plant-based options", "new_text": "Dietary preference: Vegetarian (6 months); favorite plant milk: oat milk; makes vegan pancakes for kids on Sundays"}
\end{lstlisting}

\paragraph{Analysis.} The Base model outputs unstructured fragments. SFT learns valid formatting but defaults to REWRITE, risking information loss when reorganizing the entire block. The RL model learns that REPLACE achieves targeted updates while preserving unrelated fields (Family, Hobbies).

\subsection{Episodic Memory: Temporal Chain Construction}

\paragraph{Scenario.} The user follows up on a previously mentioned trip. The retrieved memory contains:
\begin{lstlisting}
[EPISODIC] 2024-02-15: User planned trip to Japan | Details: Planning vacation to Japan, interested in visiting Kyoto and Tokyo. Researching cherry blossom season timing.
\end{lstlisting}

\paragraph{User Message.} ``Remember when I told you about my trip to Japan last month? Well, I finally uploaded all the photos! I got some amazing shots at the bamboo forest in Kyoto.''

\paragraph{Base Model.} Adds a disconnected entry with no reference to the planning phase:
\begin{lstlisting}
{"operations": [{"action": "ADD", "memory": "User went to Japan. Took photos at bamboo forest."}]}
\end{lstlisting}

\paragraph{SFT Model.} Overwrites the planning record with completion information:
\begin{lstlisting}
{"operations": [{"action": "UPDATE", "old_memory": "2024-02-15: User planned trip to Japan...", "new_memory": "2024-03-15: User completed trip to Japan | Details: Took trip to Japan last month. Visited Kyoto and Tokyo. Took photos at bamboo forest in Arashiyama. Uploaded all photos."}]}
\end{lstlisting}

\paragraph{RL Model.} Creates a new event entry and updates the original with status annotation:
\begin{lstlisting}
{"operations": [{"action": "ADD", "memory": "2024-03-10: User visited bamboo forest in Arashiyama, Kyoto | Details: Took amazing photos during trip to Japan; shared photos highlighting scenic beauty and personal enjoyment"}, {"action": "UPDATE", "old_memory": "2024-02-15: User planned trip to Japan...", "new_memory": "2024-02-15: User planned trip to Japan [COMPLETED] | Details: Planning vacation to Japan -> Trip completed March 2024. Visited Kyoto (bamboo forest) and Tokyo. Photos uploaded."}]}
\end{lstlisting}

\paragraph{Analysis.} The Base model creates isolated entries without temporal connections. SFT's UPDATE operation overwrites planning details, losing information about the user's original interests (cherry blossom season research). The RL model constructs a temporal chain: preserving the planning phase, adding the completion event as a separate entry, and marking status transitions.

\subsection{Semantic Memory: Comprehensive Fact Capture}

\paragraph{Scenario.} The user expresses interest in art after visiting a museum.

\paragraph{User Message.} ``I've been really into impressionist art lately. Went to see the Monet exhibition at the museum last weekend. His water lilies series is absolutely stunning.''

\paragraph{Base Model.} Adds a simple entry with minimal detail:
\begin{lstlisting}
{"operations": [{"action": "ADD", "memory": "User likes impressionist art. Saw Monet exhibition."}]}
\end{lstlisting}

\paragraph{SFT Model.} Updates an existing hobby entry:
\begin{lstlisting}
{"operations": [{"action": "UPDATE", "old_memory": "Hobbies: Enjoys visiting museums | Details: Regular museum visitor...", "new_memory": "Hobbies: Enjoys visiting museums | Details: Regular museum visitor, interested in various exhibitions, particularly impressionist art. Recently saw Monet's water lilies series exhibition at a museum last weekend."}]}
\end{lstlisting}

\paragraph{RL Model.} Creates multiple entries capturing both user preference and conceptual knowledge:
\begin{lstlisting}
{"operations": [{"action": "ADD", "memory": "User - Art Interest: Impressionist art | Details: Recently developed strong interest in impressionist art, particularly Monet's water lilies series. Describes them as 'absolutely stunning'."}, {"action": "ADD", "memory": "Concept - Monet's Water Lilies | Details: A series of paintings by Claude Monet depicting water lilies, displayed at Musee de l'Orangerie in Paris."}]}
\end{lstlisting}

\paragraph{Analysis.} The Base model captures only surface-level facts. SFT consolidates new information into existing entries, compressing the user's art interest into a general hobby record. The RL model learns to output multiple distinct entries: one for user-specific preferences (interest in impressionist art) and another for conceptual knowledge (Monet's Water Lilies). This separation enables more precise retrieval when answering questions about the user's artistic taste versus factual questions about specific artworks.

\subsection{Procedural Memory: Version Preservation}

\paragraph{Scenario.} The user describes a refined coffee routine. The retrieved memory contains:
\begin{lstlisting}
[PROCEDURAL] User's coffee brewing method | Steps: 1. Grind beans. 2. Add hot water. 3. Wait 4 minutes. 4. Press and pour. | Context: French press method.
\end{lstlisting}

\paragraph{User Message.} ``I finally perfected my morning coffee routine! I grind 18g of beans, use 92°C water, bloom for 30 seconds, then do a slow pour over 3 minutes.''

\paragraph{Base Model.} Adds a simple entry without procedural structure:
\begin{lstlisting}
{"operations": [{"action": "ADD", "memory": "User has new coffee method: 18g beans, 92C water, 30s bloom, 3min pour."}]}
\end{lstlisting}

\paragraph{SFT Model.} Overwrites the French press method entirely:
\begin{lstlisting}
{"operations": [{"action": "UPDATE", "old_memory": "User's coffee brewing method | Steps: 1. Grind...", "new_memory": "User's pour-over coffee brewing method | Steps: 1. Grind 18g of beans. 2. Use 92C water. 3. Bloom for 30 seconds. 4. Slow pour over 3 minutes. | Context: Pour-over method, refined morning routine."}]}
\end{lstlisting}

\paragraph{RL Model.} Adds the new method as a versioned entry, preserving the original:
\begin{lstlisting}
{"operations": [{"action": "ADD", "memory": "Pour-over coffee method (v2, current) | Steps: 1. Grind 18g beans to medium-fine. 2. Heat water to 92C. 3. Bloom for 30 seconds. 4. Slow pour in circular motion for 3 minutes. 5. Allow to drip. | Context: Refined morning routine, replaces previous French press method"}]}
\end{lstlisting}

\paragraph{Analysis.} This case exhibits the starkest contrast. SFT's UPDATE completely erases the French press procedure---if the user later asks ``How did I use to make coffee with my French press?'', the system cannot answer. The RL model learns through downstream QA feedback that procedural knowledge should be versioned rather than overwritten. The new entry explicitly references the previous method while the original remains retrievable.

\section{Additional Analysis}

\subsection{Validation of Retrieval-Based Attribution}
\label{app:retrieval_validation}

To validate that retrieval frequency serves as an adequate proxy for downstream contribution, we recruited three graduate students (M.S.\ or above in Computer Science, familiar with dialogue systems and NLP) to independently evaluate 100 synthetic QA pairs sampled from the RL training set. For each question, annotators examined whether the dominant memory type (by retrieval count) contains information directly used to derive the correct answer, with the final label determined by majority vote. Results are shown in Table~\ref{tab:retrieval_validation}.

\begin{table}[h]
\centering
\small
\resizebox{\linewidth}{!}{
\begin{tabular}{lcc}
\toprule
\textbf{Category} & \textbf{Count} & \textbf{\%} \\
\midrule
Dominant type contains answer-relevant information & 95 & 95 \\
Correct answer relies on a non-dominant type & 2 & 2 \\
Policy model failed to store relevant information & 3 & 3 \\
\bottomrule
\end{tabular}
}
\caption{Human evaluation of retrieval-based attribution on 100 sampled QA pairs.}
\label{tab:retrieval_validation}
\end{table}

Pairwise inter-annotator agreement ranges from 92.0\% to 96.0\%, with all three annotators agreeing on 92.0\% of samples. While retrieval frequency is not a perfect indicator, the dominant type aligns with the actual information source in 95\% of cases, confirming that it provides a probabilistically reliable attribution signal.

\subsection{Synthetic QA Quality Evaluation}
\label{app:qa_quality}

To assess the quality of synthetic QA pairs used for dense reward computation, we recruited three graduate students (M.S.\ or above in Computer Science, familiar with dialogue systems) to independently evaluate 100 randomly sampled QA pairs from the RL training set. Each pair was assessed on three criteria, with final labels determined by majority vote:

\begin{itemize}
    \item \textbf{Question Reasonableness}: whether the question is well-formed and answerable given the conversation context up to the current session.
    \item \textbf{Answer Correctness}: whether the ground-truth answer is factually correct based on the information present in the dialogue history.
    \item \textbf{Answer Specificity}: whether the answer is specific enough to unambiguously address the question, rather than being vague or overly general.
\end{itemize}

Results are shown in Table~\ref{tab:qa_quality}.

\begin{table}[h]
\centering
\small
\begin{tabular}{lcc}
\toprule
\textbf{Criterion} & \textbf{Pass Rate} & \textbf{Avg.\ Agreement} \\
\midrule
Question Reasonableness & 91\% & 91.3\% \\
Answer Correctness      & 92\% & 90.7\% \\
Answer Specificity      & 93\% & 94.0\% \\
\bottomrule
\end{tabular}
\caption{Human evaluation of synthetic QA quality on 100 sampled pairs. Agreement refers to average pairwise inter-annotator agreement.}
\label{tab:qa_quality}
\end{table}

Among the 8 incorrect answers (by majority vote), the primary error types are temporal calculation errors (6/8, e.g., incorrect date arithmetic or imprecise relative time references) and over-inference across sessions (2/8, where answers combine information from multiple sessions in ways not directly supported by the dialogue).

\subsection{Feasibility of Open-Source Answer Models}
\label{app:open_source}

Our evaluation pipeline uses Claude 4.5 Sonnet as the default answer model, which incurs API costs. To assess whether open-source models can serve as viable alternatives, we fix the memory bank to that constructed by Claude 4.5 Sonnet and evaluate different answer models on LoCoMo under identical retrieval settings. This setup isolates the answer model's capability by controlling for memory quality.

\begin{table}[h]
\centering
\small
\begin{tabular}{lc}
\toprule
\textbf{Answer Model} & \textbf{LoCoMo} \\
\midrule
Claude 4.5 Sonnet & 82.61 \\
GPT-4.1-mini      & 80.88 \\
Qwen3-30B-A3B     & 80.84 \\
\bottomrule
\end{tabular}
\caption{Performance of different answer models on LoCoMo using memory constructed by Claude 4.5 Sonnet, isolating the effect of answer model capability.}
\label{tab:open_source_answer}
\end{table}

Qwen3-30B-A3B achieves performance comparable to GPT-4.1-mini, demonstrating that open-source models of similar scale can serve as cost-effective alternatives for the answer component in both evaluation and RL reward computation.

\section{Prompt Templates}
\label{app:prompts}

This section presents the prompt templates used in our framework. For readability, we have made minor formatting adjustments (e.g., line breaks and indentation) to the original prompts.

\begin{figure*}[ht]
    \centering
    \includegraphics[width=\textwidth, page=1]{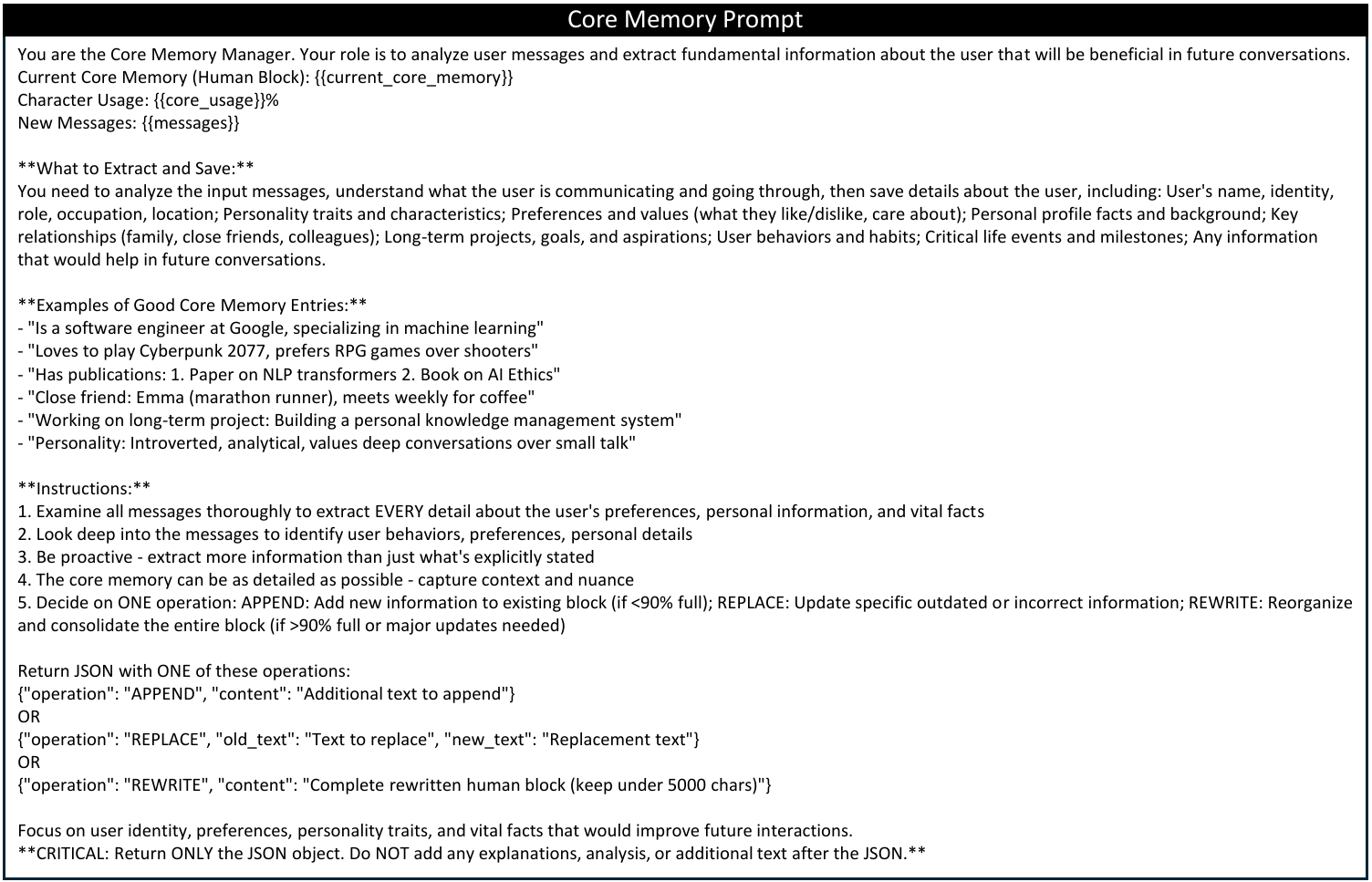}
    \caption{Prompt template for Core Memory.}
    \label{fig:prompt_core}
\end{figure*}

\begin{figure*}[ht]
    \centering
    \includegraphics[width=\textwidth, page=2]{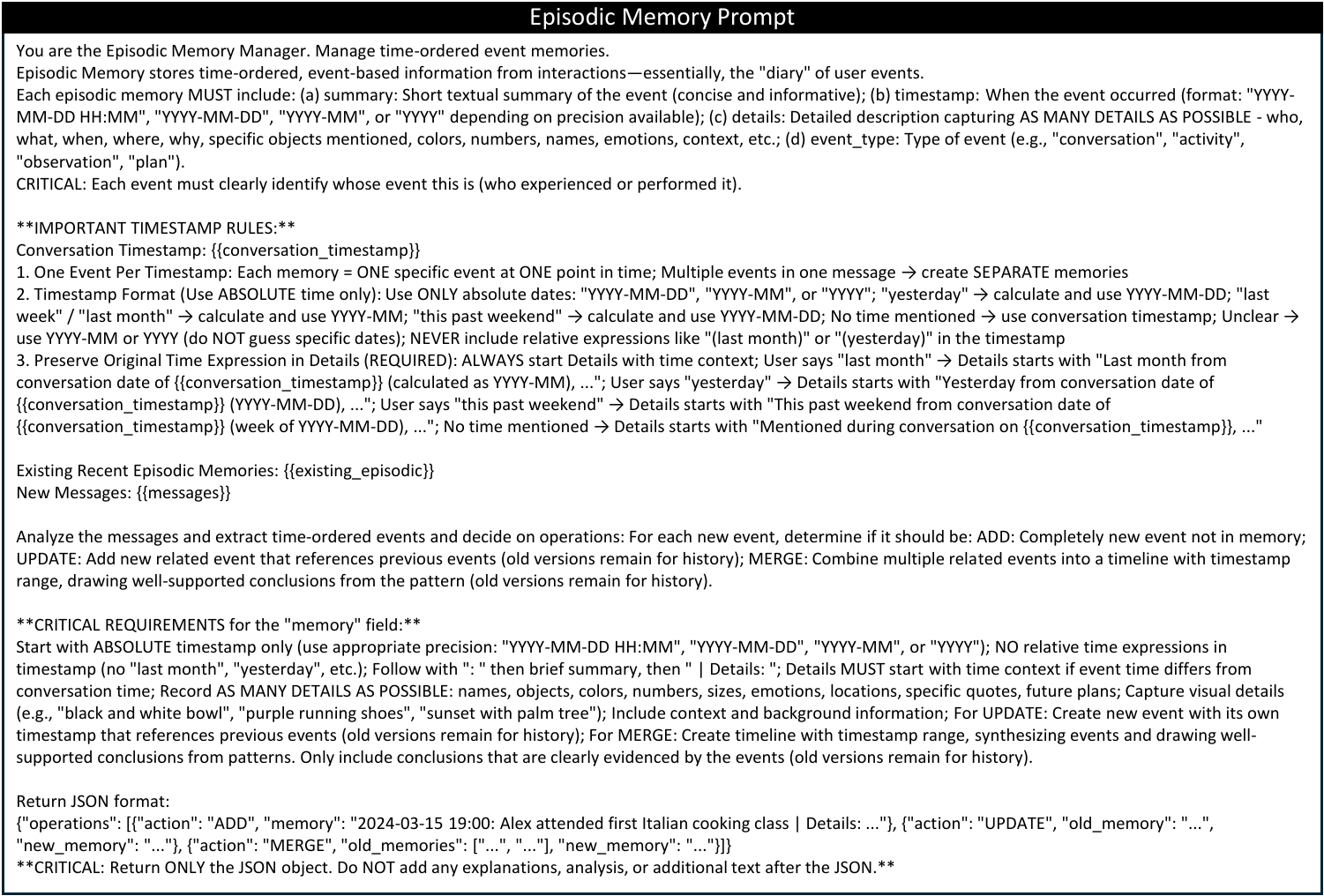}
    \caption{Prompt template for Episodic Memory.}
    \label{fig:prompt_episodic}
\end{figure*}

\begin{figure*}[ht]
    \centering
    \includegraphics[width=\textwidth, page=3]{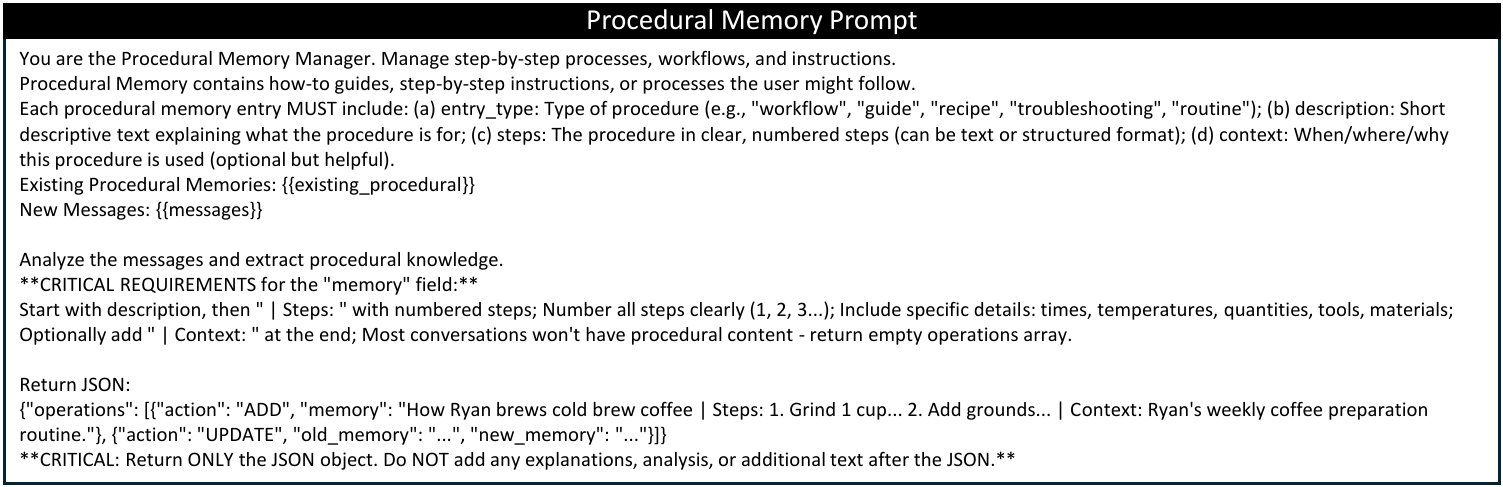}
    \caption{Prompt template for Procedural Memory.}
    \label{fig:prompt_procedural}
\end{figure*}

\begin{figure*}[ht]
    \centering
    \includegraphics[width=\textwidth, page=4]{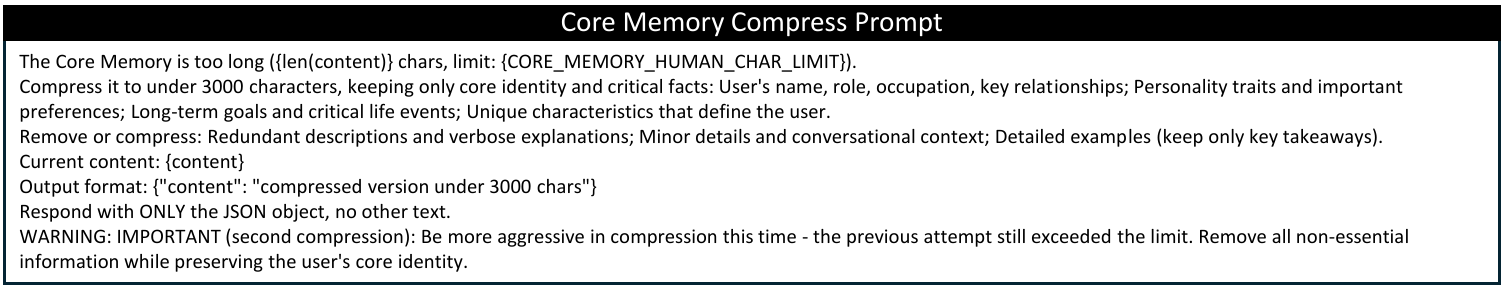}
    \caption{Prompt template for Core Memory compression.}
    \label{fig:prompt_compress}
\end{figure*}

\begin{figure*}[ht]
    \centering
    \includegraphics[width=\textwidth, page=5]{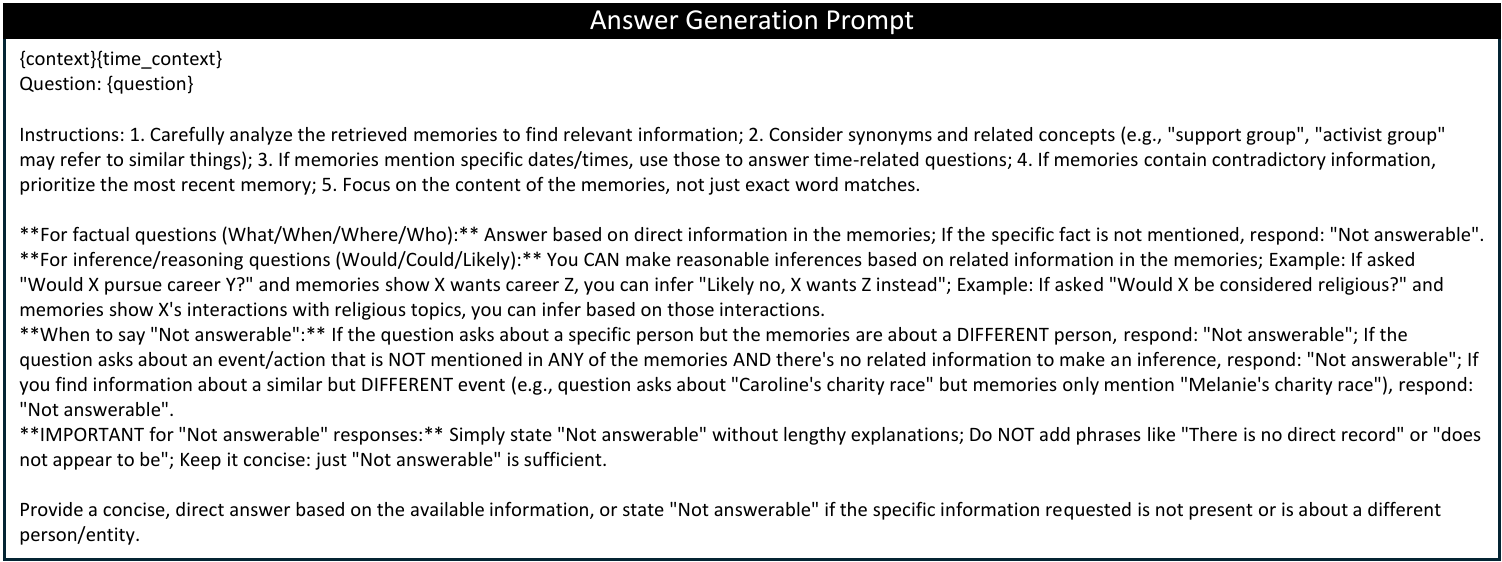}
    \caption{Prompt template for QA answering.}
    \label{fig:prompt_answer}
\end{figure*}

\begin{figure*}[ht]
    \centering
    \includegraphics[width=\textwidth, page=6]{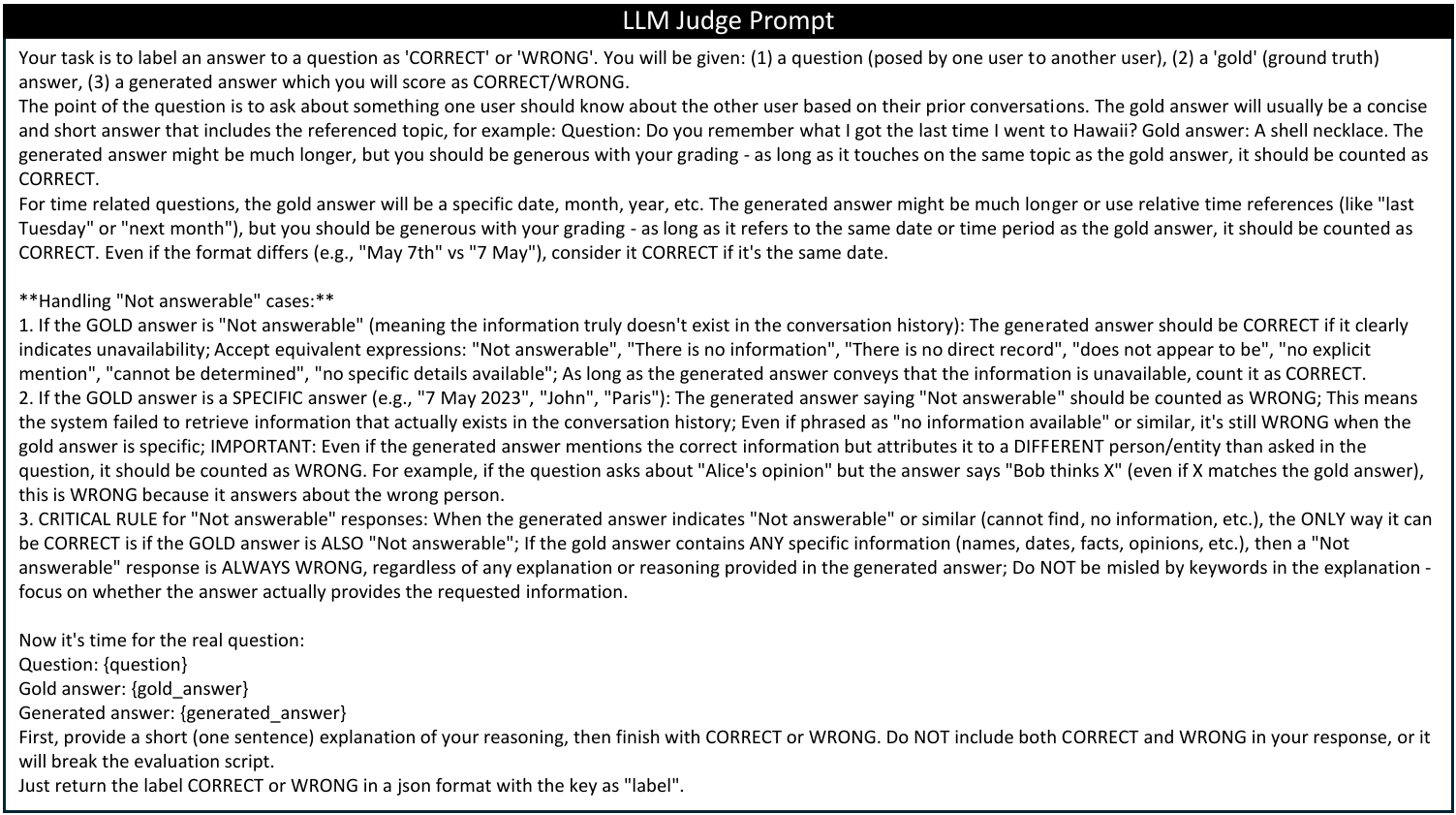}
    \caption{Prompt template for LLM judge evaluation.}
    \label{fig:prompt_judge}
\end{figure*}

\begin{figure*}[ht]
    \centering
    \includegraphics[width=\textwidth, page=7]{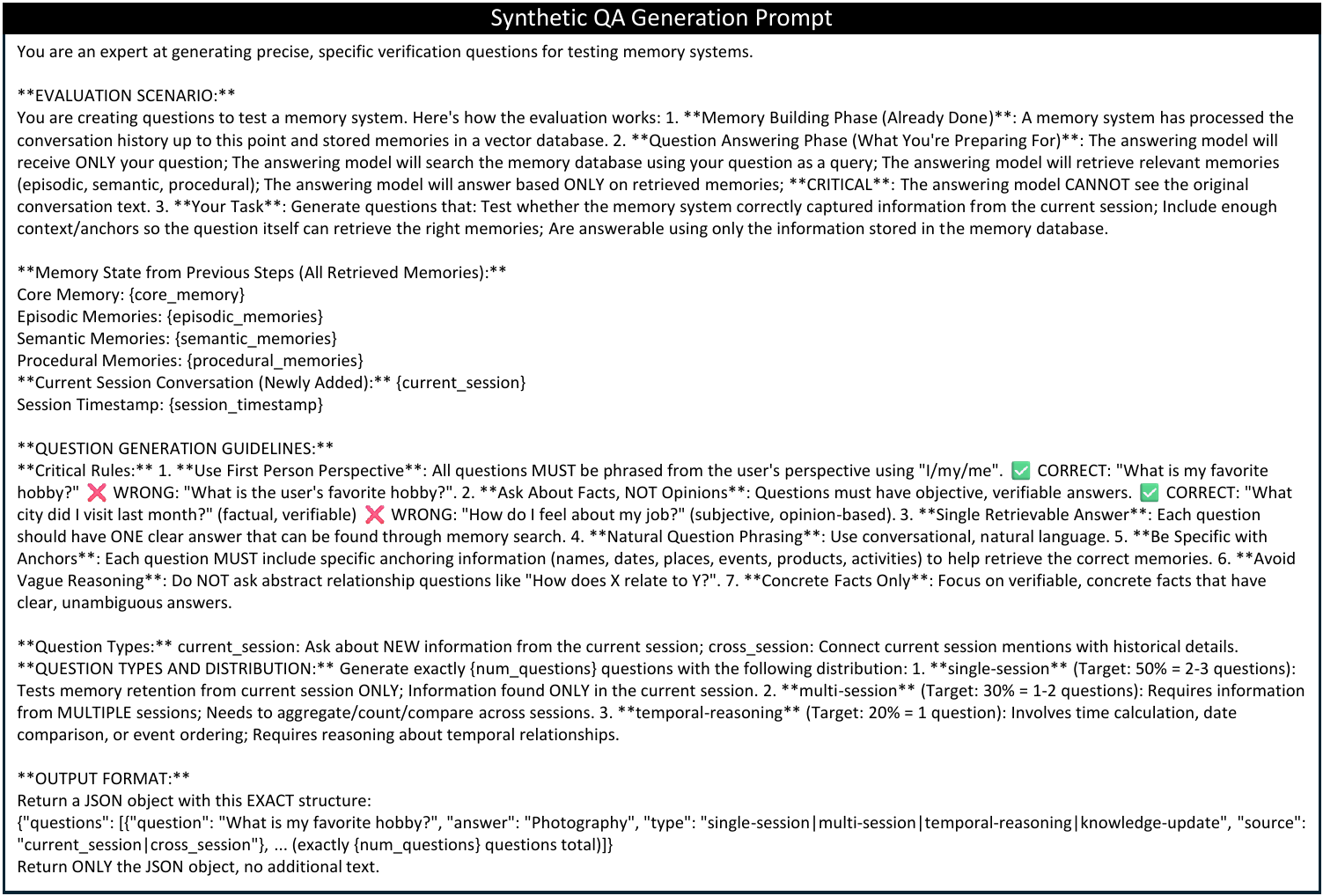}
    \caption{Prompt template for synthetic question generation.}
    \label{fig:prompt_qa_gen}
\end{figure*}

\end{document}